\documentclass[10pt,twocolumn,letterpaper]{article}

\usepackage{iccv}
\usepackage{times}
\usepackage{epsfig}
\usepackage{graphicx}
\usepackage{amsmath}
\usepackage{amssymb}

\usepackage{multirow}
\usepackage[nocompress]{cite}
\usepackage[normalem]{ulem}

\usepackage[pagebackref=true,breaklinks=true,letterpaper=true,colorlinks,bookmarks=false]{hyperref}

\iccvfinalcopy % *** Uncomment this line for the final submission

\def\iccvPaperID{} % *** Enter the ICCV Paper ID here

\ificcvfinal\fi
\begin{document}

%%%%%%%%% TITLE
\title{Transferable Semi-supervised 3D Object Detection from RGB-D Data}

\author{Yew Siang Tang \qquad Gim Hee Lee\\
Department of Computer Science, National University of Singapore\\
{\tt\small \{yewsiang.tang, gimhee.lee\}@comp.nus.edu.sg}}

\maketitle
%\thispagestyle{empty}

%%%%%%%%% ABSTRACT
\begin{abstract}
%It is well known that examples from a specific object class with ground truth labels must be shown to a fully supervised object detection network during training, for the network to detect objects from the same class during inference. Unfortunately, it is difficult to obtain 3D ground truth labels for a large number of object classes to train a fully supervised 3D object detector.
% In this paper, we learn to perform 3D object detection on new object classes with only 2D box labels to reduce the dependency on getting the comparatively more costly 3D box labels for 
% the new object classes. 
We investigate the direction of training a 3D object detector for new object classes from only 2D bounding box labels of these new classes, while simultaneously transferring information from 3D bounding box labels of the existing classes.
To this end, we propose a transferable semi-supervised 3D object detection model that learns a 3D object detector network from training data with two disjoint sets of object classes - a set of strong classes with both 2D and 3D box labels, and another set of weak classes with only 2D box labels. In particular, we suggest a relaxed reprojection loss, box prior loss and a Box-to-Point Cloud Fit network that allow us to effectively transfer useful 3D information from the strong classes to the weak classes during training, and consequently, enable the network to detect 3D objects in the weak classes during inference. 
{\color{black} Experimental results show that our proposed algorithm outperforms baseline approaches and achieves promising results compared to fully-supervised approaches on the SUN-RGBD and KITTI datasets. Furthermore, we show that our Box-to-Point Cloud Fit network improves performances of the fully-supervised approaches on both datasets. }
\end{abstract}

%%%%%%%%% BODY TEXT
\vspace{-0.15in}
\section{Introduction}

\begin{figure}[t]
\begin{center}
\includegraphics[width=\linewidth]{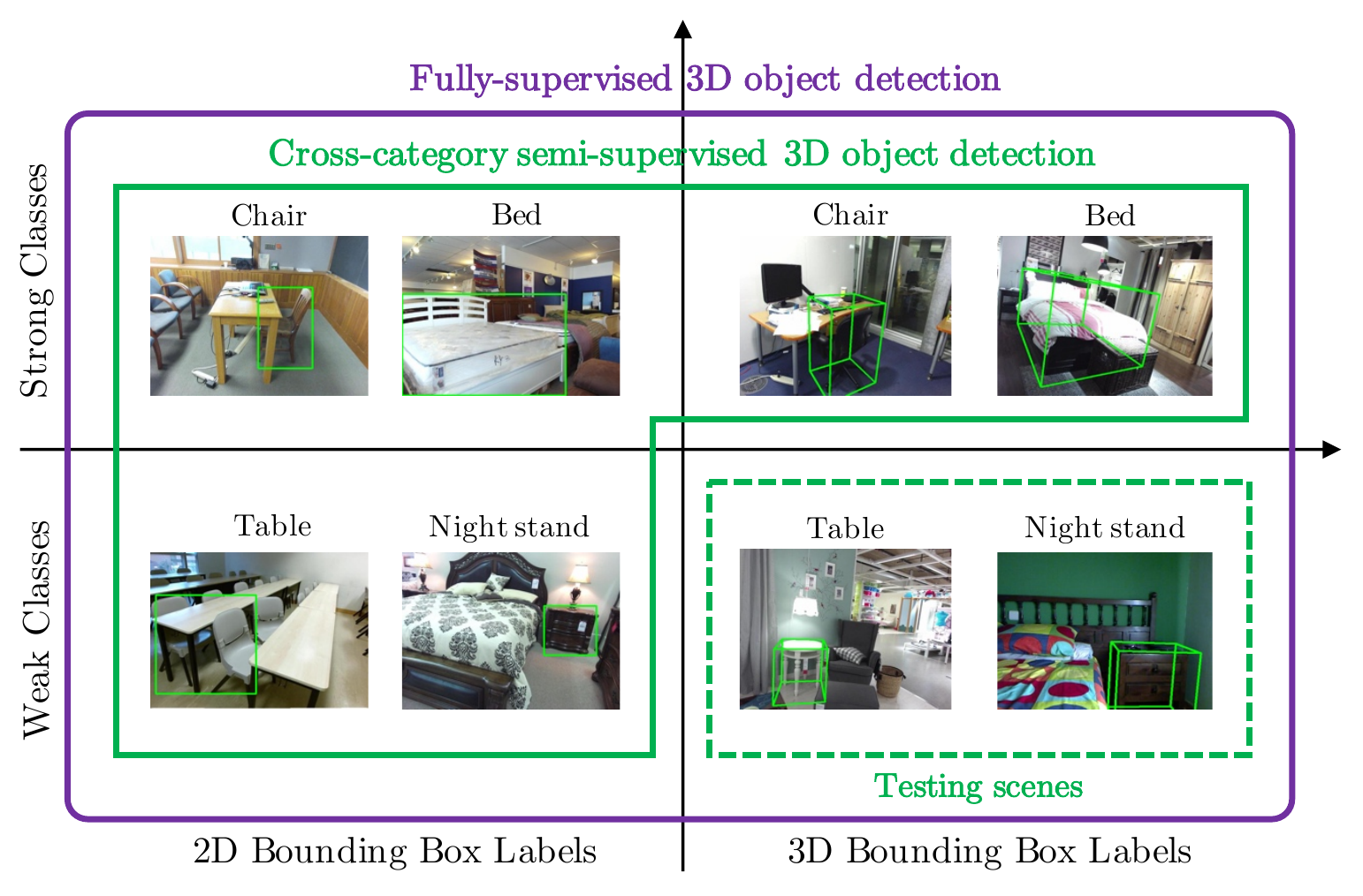}
\end{center}
\vspace{-0.05in}
\caption{Illustration of fully-supervised 3D object detection where 3D box labels of all classes (\textit{weak} and \textit{strong} classes) are available, and cross-category semi-supervised 3D object detection where 3D box labels of inference classes (\textit{weak} classes e.g. \texttt{Table} and \texttt{Night stand}) are not available. 
Conventional semi-supervised learning (which is in-category) requires strong annotations from the inference classes (3D box labels in the case of 3D object detection) while cross-category semi-supervised learning requires only the weak annotations from the inference classes (2D box labels in the case of 3D object detection).
}
\label{fig:teaser}
\vspace{-0.15in}
\end{figure}

\begin{figure*}[t]
\begin{center}
\includegraphics[width=0.98\linewidth]{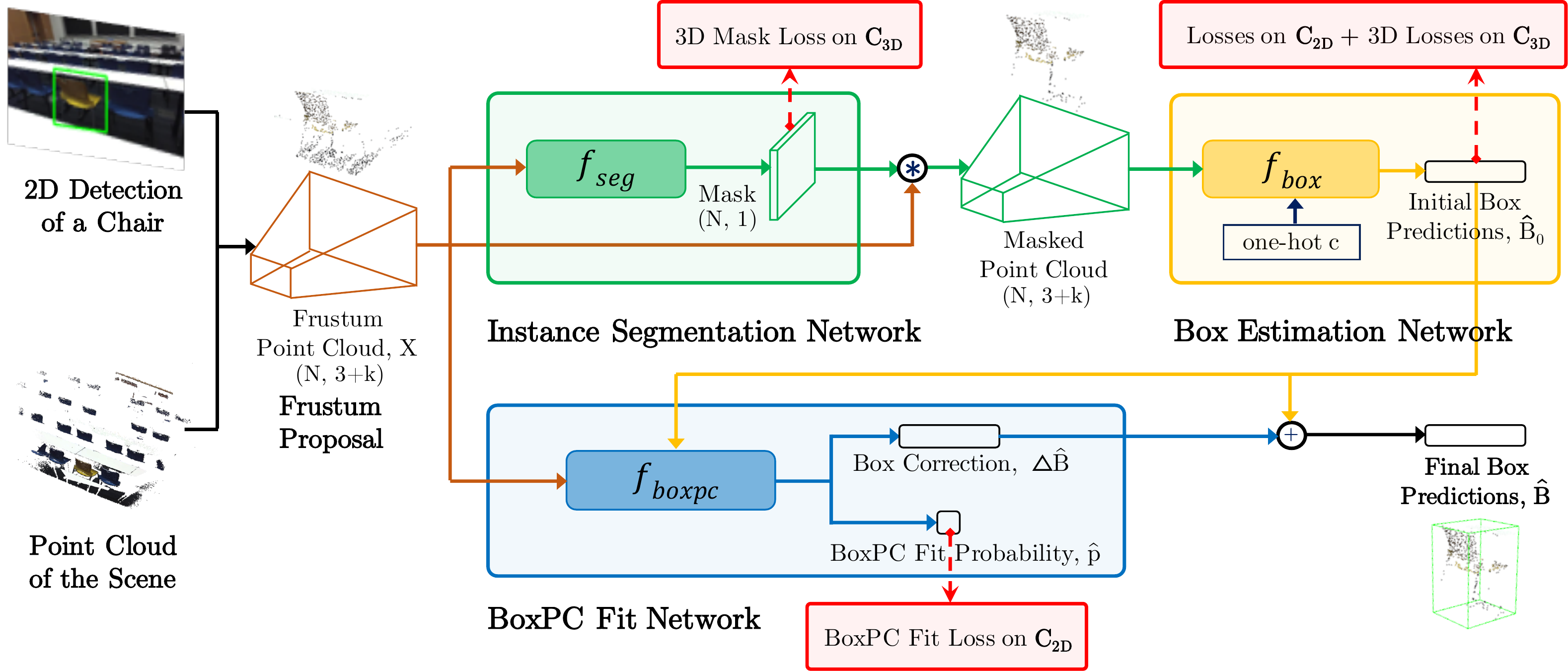}
\end{center}
\caption{ Overall architecture with a RGB-D (a pair of image and point cloud) input. (1) 
Frustum point clouds $X$ are extracted from the input point cloud and 2D object detection boxes on the image. 
(2) 
$f_{seg}$ takes $X$ as input and outputs class-agnostic segmentations used to mask $X$. 
(3) $f_{box}$ predicts an initial 3D box $\hat{B}_0$ with the masked point cloud. 
% (2) $X$ is fed into $f_{seg}$ to produce class-agnostic segmentations.
% (3) $X$ is masked by the segmentations before $f_{box}$ uses it to predict an initial 3D box $\hat{B}_0$.
% (4) The pretrained $f_{boxpc}$ model takes in $X$ and $\hat{B}_0$, and predicts a refinement term and a BoxPC fit probability term used to supervise $f_{box}$.
(4) The pretrained $f_{boxpc}$ model refines $\hat{B}_0$ to $\hat{B}$ according to $X$, and predicts the BoxPC fit probability used to supervise $f_{box}$.}
\label{fig:MainNetwork}
\vspace{-0.05in}
\end{figure*}

3D object detection refers to the problem of classification and estimation of a tight oriented 3D bounding box for each object in a 3D scene represented by a point cloud. It plays an essential role in many real-world applications such as home robotics, augmented reality and autonomous driving, where machines must actively interact and engage with its surroundings. In the recent years, significant advances in 3D object detection have been achieved with fully supervised deep learning-based approaches \cite{Vote3D, 3DOP, Mono3D, MV3D, Deep3DBox, VoxelNet, PointFusion, AVOD, FPN}. 
% However, it is well known that training data from a specific object class must be presented to the deep network during training for it to successfully learn to detect the same object class during inference. It is often tedious, time consuming and labor intensive to manually label bounding boxes in the 3D point cloud. 
% The problem is further aggravated due to the need to label large amounts of 3D bounding boxes over a huge variety of object classes. 
% Consequently, the existing 3D datasets contain limited number of labels for many object classes except for a few dominant ones \cite{SUN,Kitti,ObjectNet3d,Pascal3D}. This makes it difficult for new applications to go beyond the available classes in these public datasets.
% On the contrary, it is much easier to obtain and train on a large number of 2D bounding box labels in a wide variety of object classes \cite{PascalVOC,COCO,Imagenet,SUN,Kitti,ObjectNet3d,Pascal3D}. 
% {\color{blue} Calculations based on \cite{ExtremeClicking} show that it is $\sim$3-16 times faster to label 2D than 3D bounding boxes (details in supplementary). }
{\color{black} However, these strongly supervised approaches rely on 3D ground truth datasets \cite{SUN,Kitti,ObjectNet3d,Pascal3D} that are tedious and time consuming to label even with good customised 3D annotation tools such as those in \cite{SUN,Kitti}. In contrast, calculations based on \cite{ExtremeClicking} show that it can be $\sim$3-16 times faster (depending on the annotation tool used) to label 2D than 3D bounding boxes (details in supplementary). This observation inspires us to investigate the direction of training a 3D object detector for new object classes from only 2D bounding box labels of these new classes, while simultaneously transferring information from 3D bounding box labels of the existing classes. We hope that this direction of research leads to cost reductions, and improves the efficiency and practicality of learning 3D object detectors in new applications with new classes. }

More specifically,
the objective of this paper is to train a 3D object detector network with data from two disjoint sets of object classes - a set of classes with both 2D and 3D box labels (\textit{strong} classes), and another set of classes with only 2D box labels (\textit{weak} classes).
In other words, our goal is to transfer useful 3D information from the \textit{strong} classes to the \textit{weak} classes {\color{black} (the new classes of a new 3D object detection application correspond to these \textit{weak} classes) }.
We refer to this as cross-category semi-supervised 3D object detection as illustrated in Fig.~\ref{fig:teaser}. 

To this end, we propose a novel transferable semi-supervised 3D object detection network. Our network leverages on the state-of-the-art Frustum PointNets \cite{FPN} as the backbone. We train the backbone network to make class-agnostic segmentations and class-conditioned initial 3D box predictions on the \textit{strong} classes in a fully-supervised manner. We also supervise the network with 2D box labels of the \textit{weak} classes using a relaxed reprojection loss which corrects 3D box predictions that violate boundaries specified by the 2D box labels and through the prior knowledge of the object sizes. To transfer knowledge from 3D box labels from the \textit{strong} classes, we first train a Box-to-Point Cloud (BoxPC) Fit network on 3D box and point cloud pairs to reason about 3D boxes and the fit with their corresponding point clouds. More specifically, we proposed an effective and differentiable method to encode a 3D box and point cloud pair. Finally, the 3D box predictions on the \textit{weak} classes are supervised and refined by the BoxPC Fit network. \\

% Contributions of paper
\noindent The contributions of our paper are as follows:
\begin{itemize}
\vspace{-0.05in}
\item We propose a network to perform 3D object detection on \textit{weak} classes, where \textit{only} 2D box labels are available. This is achieved using relaxed reprojection and box prior losses on the \textit{weak} classes, and transferred knowledge from the \textit{strong} classes.
\item A differentiable BoxPC Fit network is designed to effectively combine BoxPC representation. It is able to supervise and improve the 3D object detector on the \textit{weak} classes after training on the \textit{strong} classes.
\vspace{-0.05in}
\item {\color{black} Our transferable semi-supervised 3D object detection model outperforms baseline approaches and achieves promising results compared to fully-supervised methods on both SUN-RGBD and KITTI datasets. Additionally, we show that our BoxPC Fit network can be used to improve the performance of fully-supervised 3D object detectors. }
% \item Our transferable semi-supervised 3D object detection model achieves strong performance that greatly exceeds baseline approaches and is competitive with fully-supervised methods on both SUN-RGBD and KITTI 3D object detection datasets. Additionally, we show the usefulness of our BoxPC Fit network in fully-supervised settings by improving the performance of fully-supervised 3D object detectors significantly.
\end{itemize}

\section{Related Work}

%-------------------------------------------------------------------------
\paragraph{3D Object Detection} 3D object detection approaches have advanced significantly in the recent years \cite{LSS, COG, 2DDriven3D, Vote3D, 3DOP, Mono3D, MV3D, Deep3DBox, VoxelNet, PointFusion, AVOD, FPN}. However, most approaches are fully-supervised and highly-dependent on 3D box labels that are diffcult to obtain. To the best of our knowledge, there are no existing weakly- or semi-supervised 3D object detection approaches.
Recently, \cite{Implicit} proposed a self-supervised Augmented Autoencoder that trains on CAD models, removing the need for pose-annotated training data. However, it is unclear if the method generalizes to general 3D object detection datasets with high intra-class shape variations and noisy depth data. 

\vspace{-0.05in}
\paragraph{Weakly- and Semi-Supervised Learning}
There is growing interest in weakly- and semi-supervised learning in many problem areas \cite{3DShapeModelsFrom2DImages, 3DHumanPoseWeak, NetVLAD, BeatMTurkers, Weakly3DHPE, 3DFeatNet, WeakRPN, WeakShapeCompletionFromLaser} because it is tedious and labor intensive to label large amounts of data for fully supervised deep learning. 
There is a wide literature of weakly-supervised \cite{SEC,CCNN,RandomWalk,WeakInstSegClassPeak} and semi-supervised \cite{WeakAndSemiSemSeg,LearningToSegUnderVariousWeak,Decoupled,WeakSemiPanop} learning approaches for semantic segmentation. Both strong and weak labels on the inference classes are required in conventional semi-supervised learning. Hence, the approach remains expensive for applications with new classes. \cite{Trans1,Trans2,Trans3} proposed the cross-category semi-supervised semantic segmentation which is a more general approach of semi-supervised segmentation, where the model is able to learn from strong labels provided for classes outside of the inference classes.
They outperformed weakly- and semi-supervised methods, 
and showed on-par performances with fully-supervised methods. Inspired by the effectiveness of the transferred knowledge, we propose to tackle the same problem in the 3D object detection domain.

\vspace{-0.025in}
\section{Problem Formulation}

Let $C = C_{2D} \cup C_{3D}$, where $C_{2D}$ is the set of classes with \textit{only} 2D box labels (\textit{weak} classes) and $C_{3D}$ is the disjoint set of classes with 2D and 3D box labels (\textit{strong} classes). We tackle the cross-category semi-supervised 3D object detection problem where 3D object detection is performed on the object classes in $C_{2D}$ while training is done on strong labels from $C_{3D}$ and weak labels from $C_{2D}$. In 3D object detection with RGB-D data, our goal is to classify and predict \textit{amodal} 3D bounding boxes for objects in the 3D space. Depth data can be provided by various depth sensors e.g. RGB-D sensors and LiDAR etc. Each 3D bounding box is parameterized by its center $(b_x,b_y,b_z)$, size $(b_h,b_w,b_l)$ and orientation $b_{\theta}$ along the vertical axis.

\vspace{-0.025in}
\section{Method}

\subsection{Frustum PointNets} \label{sec:fpn}
In this section, we briefly discuss Frustum PointNets (FPN), the 3D object detection framework which we adopt as the backbone of our framework. Refer to \cite{FPN} and our supplementary for more details.

\vspace{-0.15in}
\subsubsection{Frustum Proposal}
As seen in Fig.~\ref{fig:MainNetwork}, the inputs to the network is an image with a 2D box for each object and a 3D point cloud. During inference, we obtain the 2D boxes from a 2D object detector trained on $C$. 
We project the point cloud onto the image plane using the camera projection matrix and select only the points that lie in the 2D box for each object. This reduces the search space to only points within a 3D frustum which we refer to as frustum point cloud. Next, variance of the points in each frustum point cloud is reduced by rotating the frustum about the vertical axis of the camera to face the front. Formally, let us denote a frustum point cloud by $X = [x_1, x_2, ..., x_N]^T\in \mathbb{R}^{N \times (3+k)}$, where $N$ is the number of points. $x_n \in \mathbb{R}^{3+k}$  (i.e. x, y, z, RGB or reflectance) is the $n$-th point in the point cloud.

\vspace{-0.1in}
\subsubsection{3D Instance Segmentation}
The frustum point cloud contains foreground points of an object which we are interested in and background points from the surroundings. To isolate the foreground object and simplify the subsequent 3D box estimation task, the frustum point cloud is fed into a 3D instance segmentation network $f_{seg}$ which predicts a class-agnostic foreground probability for each point. These probabilities are used to mask and retrieve the points with high values, thus giving a masked point cloud. $f_{seg}$ is trained by minimizing 
\begin{align}
L_{seg}^{3D} &= \sum_{c \in C_{3D}} H (f_{seg}^{*}(X^c), f_{seg}(X^c)) \;
\label{eq:Loss_s}
\end{align}
on $C_{3D}$, where $H$ is the point-wise binary cross-entropy loss and $f^*_{seg}$ is the ground truth mask. 

\begin{figure}[t]
\begin{center}
\includegraphics[width=\linewidth]{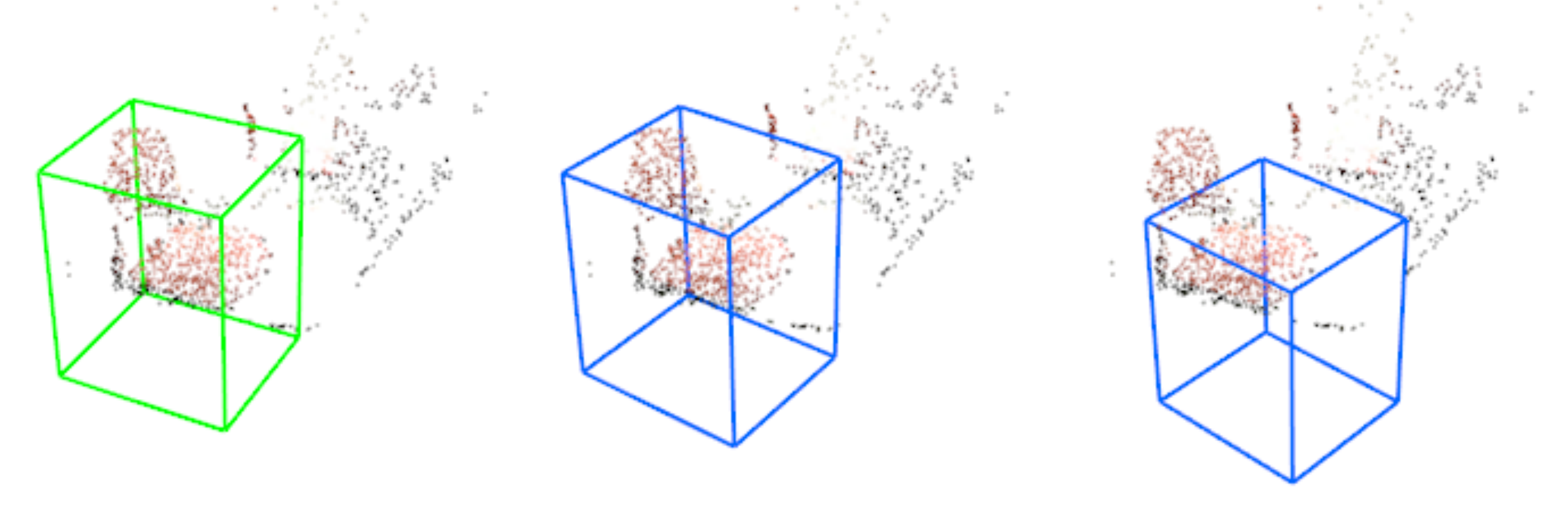}
\end{center}
   \caption{Illustration of 3D bounding box and point cloud pairs with a good BoxPC fit (center) and a bad BoxPC fit (right). The ground truth 3D bounding box of the chair is given on the left.}
\label{fig:boxpc_fit}
\vspace{-0.1in}
\end{figure}

\begin{figure*}[t]
\begin{center}
\includegraphics[width=\linewidth]{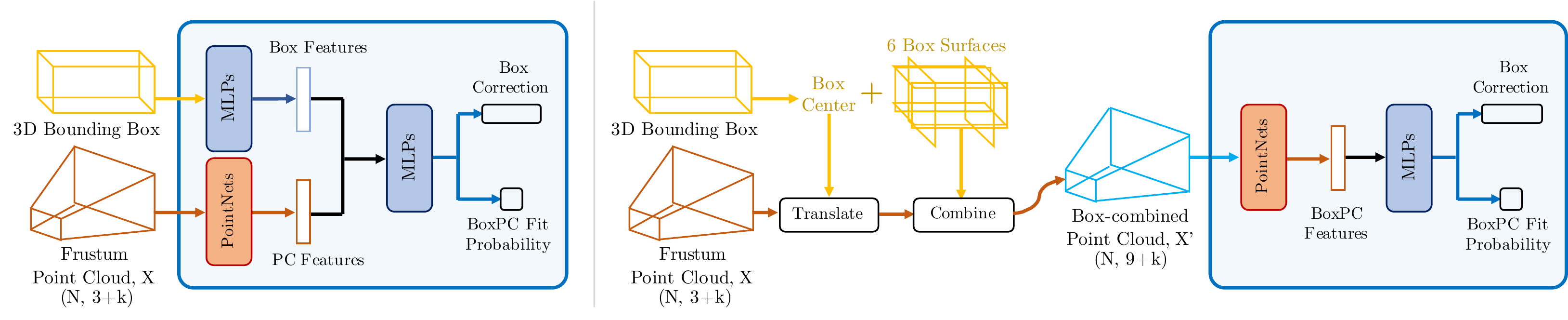}
\end{center}
\caption{ Different feature representations for a pair of 3D bounding box and point cloud. On the left, the inputs are processed independently and concatenated in the feature space. On the right, the inputs are combined in the input space and processed as a whole. }
\label{fig:network_boxpc}
\vspace{-0.1in}
\end{figure*}

\vspace{-0.1in}
\subsubsection{3D Box Estimation}
The 3D box estimation network $f_{box}$ predicts an initial 3D box of the object $\hat{B}_0$ given the masked point cloud and a one hot class vector from the 2D detector. 
Specifically, we compute the centroid of the masked point cloud and use it to translate the masked point cloud to the origin position to reduce translational variance. Next, the translated point cloud is given to two PointNet-based \cite{PN} networks that will predict the initial 3D box $\hat{B}_0$. 
The center $(b_x,b_y,b_z)$ is directly regressed by the network while an anchor-based approach is used to predict the size $(b_h,b_w,b_l)$ and rotation $b_{\theta}$. There are $NS$ size and $NH$ rotation anchors, and $3 {NS}$ size and $NH$ rotation residuals. Hence, the network has $3 + 4 NS + 2 NH$ outputs that are trained by minimizing
\vspace{0.05in}
\begin{align}
  L_{box}^{3D} &= \sum_{c \in C_{3D}} L_{c1-reg}^c + L_{c2-reg}^c + L_{r-cls}^c + L_{r-reg}^c \nonumber \\
  & \qquad\ \ \ +  L_{s-cls}^c + L_{s-reg}^c + L_{corner}^c \;
  \label{eq:L_box3d}
\end{align}
on $C_{3D}$.
{\color{black}
$L_{c1-reg}^c$ and $L_{c2-reg}^c$ are regression losses for the box center predictions, $L_{r-cls}^c$ and $L_{r-reg}^c$ are the rotation anchor classification and regression losses, $L_{s-cls}^c$ and $L_{s-reg}^c$ are the size anchor classification and regression losses, and $L_{corner}^c$ is the regression loss for the 8 vertices of the 3D box.
Note that each loss term in $L_{box}^{3D}$ is weighted by $w_s$, which we omit in Eq. \ref{eq:L_box3d} and all subsequent losses for brevity.
Until this point, the network is fully-supervised by 3D box labels available only for classes $C_{3D}$. In the subsequent sections, we describe our contributions to train $f_{box}$ on the \textit{weak} classes in $C_{2D}$ by leveraging on the BoxPC Fit Network (Sec.~\ref{sec:boxpcfitnet}) and weak losses (Sec.~\ref{sec:weak_losses}). 
}
% We note that until this point, the network is fully-supervised by 3D box labels available only for classes $C_{3D}$. In the subsequent sections, we describe our novel method to train $f_{box}$ on the \textit{weak} classes in $C_{2D}$. 
% It is important to note that each loss term in $L_{box}^{3D}$ is weighted by $w_s$, which we omit in Eq. \ref{eq:L_box3d} and all subsequent losses for brevity.

\subsection{Box to Point Cloud (BoxPC) Fit Network} \label{sec:boxpcfitnet}
The initial 3D box predictions $\hat{B}_0$ on classes $C_{2D}$ are not reliable because there are no 3D box labels for these classes. As a result, the 3D box surfaces of the initial predictions $\hat{B}_0$ for classes $C_{2D}$ are likely to cut the frustum point cloud $X$ at unnatural places, as illustrated on the right of Fig.~\ref{fig:boxpc_fit}.
%since it does not know how the object of interest looks like or should be fitted with a 3D box. 
We utilize our BoxPC Fit Network $f_{boxpc}$ with a novel training method to transfer the knowledge of a good BoxPC fit from the 3D box labels of classes $C_{3D}$ to classes $C_{2D}$. The input to our $f_{boxpc}$ is a pair of 3D box $B$ and frustum point cloud $X$. The outputs of $f_{boxpc}$ are the BoxPC Fit probability, i.e. goodness-of-fit
\begin{equation}
    \hat{p} = f_{boxpc-cls}(X,B) \in [0,1],
\end{equation}
between $B$ and the object in $X$, and the correction
\begin{equation}
    \Delta \hat{B} = f_{boxpc-reg}(X,B) \in \mathbb{R}^7,
\end{equation}
required on $B$ to improve the fit between $B$ and the object in $X$. $\hat{p} \rightarrow 1$ when $B$ encloses the object in $X$ tightly, i.e. there is a high overlap between $B$ and the 3D box label $B^*$.

A pretrained $f_{boxpc}$ (see Sec.~\ref{train_boxpc} for the training procedure) is used to supervise and improve 3D box predictions $\hat{B}$ for classes $C_{2D}$. Specifically, we train $f_{box}$ to make better initial 3D box predictions $\hat{B}_0$ by maximizing $\hat{p}$. By minimizing the loss:
\begin{align}
  L_{fit}^{2D} &= \sum_{c \in C_{2D}} -\log(f_{boxpc-cls}(X^c, \hat{B}^c_0)),
  \label{eq:Loss_fit2d}
\end{align}
i.e. maximizing $\hat{p}$, our network $f_{box}$ learns to predict $\hat{B}_0$ that fit objects in their respective point clouds well. Finally, we obtain the final 3D box prediction $\hat{B}$ by correcting the initial 3D box prediction $\hat{B}_0$ with $\Delta\hat{B}$, i.e. $\hat{B} = \hat{B}_0 + \Delta \hat{B}$.

\subsubsection{Pretraining BoxPC Fit Network} \label{train_boxpc}
We train the BoxPC Fit network $f_{boxpc}$ on the classes in $C_{3D}$ with the 3D box labels $B^* \in \mathbb{R}^7$ and their corresponding point clouds $X$. 
We sample varying degree of perturbations $\delta \in \mathbb{R}^7$ to the 3D box labels and get perturbed 3D boxes $B^* - \delta$. 
We define 2 sets of perturbations, $\textbf{P}^+(B^*)$ and $\textbf{P}^-(B^*)$, where $\textbf{P}^+$ and $\textbf{P}^-$ are sets of small and large perturbations, respectively. Formally, $\textbf{P}^+(B^*) = \{ \delta : {\alpha}^+ \leq IOU(B^* - \delta, B^*) \leq {\beta}^+ \mid \{ {\alpha}^+, {\beta}^+ \} \in [0,1] \}$, where $IOU$ refers to the 3D Intersection-over-Union. $\textbf{P}^-$ is defined similarly but with ${\alpha}^-$ and ${\beta}^-$ as the bounds. ${\alpha}^+$ and ${\beta}^+$ are set such that the 3D boxes perturbed by $\delta \in \textbf{P}^+$ have high overlaps with $B^*$ or good fit with its point clouds and vice versa for $\textbf{P}^-$.
The loss function to train $f_{boxpc}$ is given by
\begin{equation}
    L_{boxpc}^{3D} = \sum_{c \in C_{3D}} L_{cls}^c + L_{reg}^c,
    \label{eq:L_boxpc3d}
\end{equation}  
where $L_{cls}^c$ is the classification loss
\begin{equation}
L_{cls} =
\begin{cases}
H(1,f_{boxpc-cls}(X, B^* - \delta)), &\text{if $\delta \in \textbf{P}^+ $},\\
H(0,f_{boxpc-cls}(X, B^* - \delta)), &\text{if $\delta \in \textbf{P}^- $},
\end{cases} \\
\end{equation}
to predict the fit of a perturbed 3D box and its frustum point cloud, and 
$L_{reg}$ is the regression loss 
\begin{equation}
L_{reg} = l(\delta, f_{boxpc-reg}(X, B^* - \delta)), 
\end{equation}
to predict the perturbation to correct the perturbed 3D box. $H$ is the binary cross-entropy and $l(y^*,\hat{y})$ is the Smooth L1 loss with $y^*$ as the target for $\hat{y}$.

\vspace{-0.1in}
\subsubsection{BoxPC Feature Representations}
A possible way to encode the feature respresentations for the BoxPC network is illustrated on the left of Fig.~\ref{fig:network_boxpc}.
The features of the input 3D box and frustum point cloud are learned using several MLPs and PointNets \cite{PN}, respectively. These features are then concatenated and fed into several layers of MLPs to get the BoxPC fit probability $\hat{p}$ and Box Correction term $\Delta \hat{B}$.
However, this neglects the intricate information of how the 3D box surfaces cut the point cloud, leading to poor performance as shown in our ablation studies in Sec.~\ref{exp_ablation}. Hence, we propose another method to combine the 3D box and point cloud in the input space in a differentiable manner that also exploits the relationship between them. 
Specifically, we use the box center $(b_x,b_y,b_z)$ to translate the point cloud to the origin to enhance translational invariance to the center positions. 
Next, we use the box sizes $(b_h,b_w,b_l)$ and rotation $b_{\theta}$ to form six planes representing each of the six box surfaces. We compute a $N \times 6$ feature vector of
the perpendicular distance (with direction) from each point to each of the six planes.
This feature vector is concatenated with the original frustum point cloud $X \in \mathbb{R}^{N\times(3+k)}$ to give the box-combined frustum point cloud $X' \in \mathbb{R}^{N\times(9+k)}$ as illustrated on the right of Fig.~\ref{fig:network_boxpc}. This representation allows the network to easily reason about regions where the box surfaces cut the point cloud since the perpendicular distance will be close to 0. Additionally, it is possible to perform feature learning on the 3D box and the point cloud \textit{jointly} with a single PointNet-based \cite{PN} network, improving performance. 

\subsection{Weak Losses} \label{sec:weak_losses}
We supervise the initial 3D box predictions $\hat{B}_0$ with 2D box labels $B^*_{2D}$ and additional priors. 
\vspace{-0.05in}

\begin{figure}[t]
\begin{center}
\includegraphics[width=\linewidth]{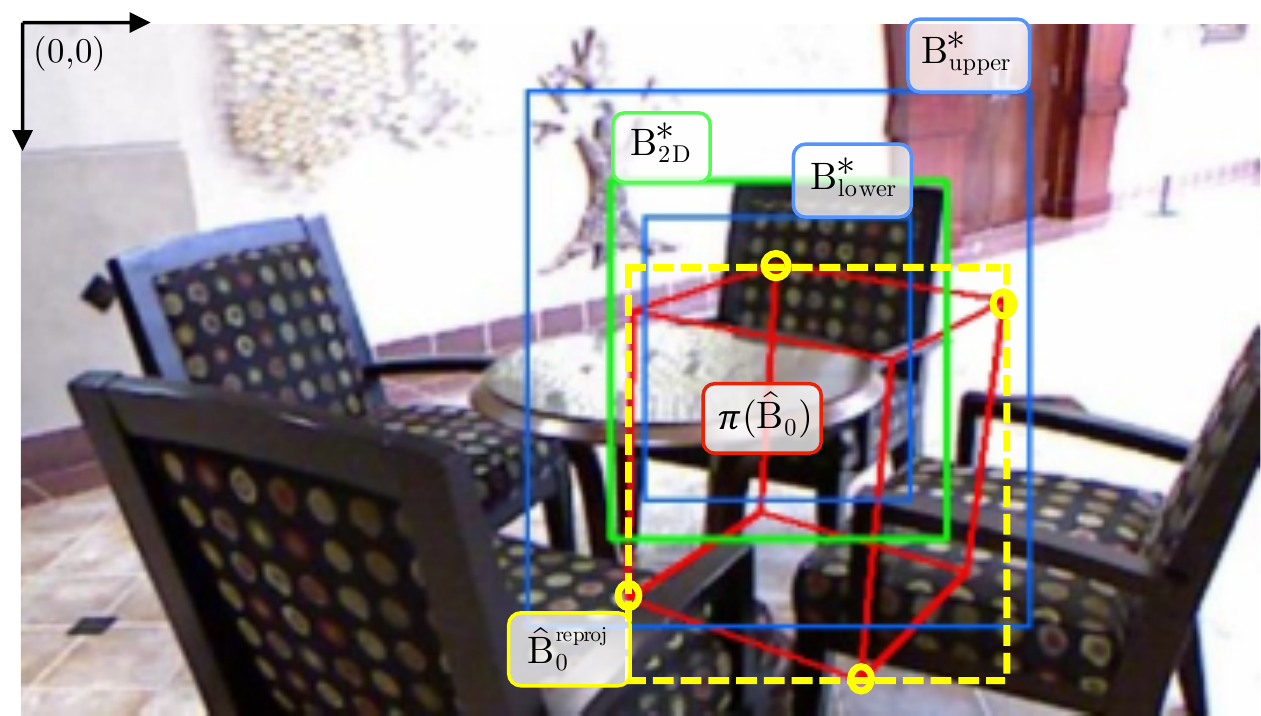}
\end{center}
   \caption{Illustration of the relaxed reprojection loss. }
\label{fig:reproj_loss}
\vspace{-0.1in}
\end{figure}

\begin{table*}
\begin{center}
\begin{tabular}{l|c c c c c c c c c c|c}
\hline
 & bathtub & bed & toilet & chair & desk & dresser & nstand & sofa & table & bkshelf & mAP \\
\hline
\textbf{Fully-Supervised:} &  &  &  &  &  &  &  &  &  &  & \\
DSS \cite{DeepSlidingShapes} & 44.2 & 78.8 & 78.9 & 61.2 & 20.5 &  6.4 & 15.4 & 53.5 & 50.3 & 11.9 & 42.1 \\
COG \cite{COG}               & 58.3 & 63.7 & 70.1 & 62.2 & \textbf{45.2} & 15.5 & 27.4 & 51.0 & 51.3 & 31.8 & 47.6 \\
2D-driven \cite{2DDriven3D}  & 43.5 & 64.5 & 80.4 & 48.3 & 27.9 & 25.9 & 41.9 & 50.4 & 37.0 & 31.4 & 45.1 \\
LSS \cite{LSS}               & \textbf{76.2} & 73.2 & 73.7 & 60.5 & 34.5 & 13.5 & 30.4 & 60.4 & \textbf{55.4} & 32.9 & 51.0 \\
FPN \cite{FPN}               & 43.3 & \textbf{81.1} & \textbf{90.9} & \textbf{64.2} & 24.7 & 32.0 & \textbf{58.1} & \textbf{61.1} & 51.1 & 33.3 & \textbf{54.0} \\
FPN*                         & 46.0 & 80.3 & 82.8 & 58.5 & 24.2 & 36.9 & 47.6 & 54.9 & 42.0 & \textbf{41.6} & 51.5 \\
FPN* + BoxPC Refine          & 51.5 & 81.0 & 85.0 & 59.0 & 24.5 & \textbf{38.6} & 52.2 & 55.3 & 43.8 & 40.9 & 53.2 \\
\hline
\textbf{CC Semi-Supervised:} &  &  &  &  &  &  &  &  &  &  & \\
FPN*                  & 6.1  & 69.5 & 28.1 & 12.4 & 18.1 & 17.5 & 37.4 & 25.8 & 23.5 & 6.9  & 24.5 \\
FPN* w/o OneHot       & 24.4 & 69.5 & 30.5 & 15.9 & 22.6 & 19.4 & 39.0 & 37.3 & 29.0 & 12.8 & 30.1 \\
Ours + R              & 29.5 & 60.9 & 65.3 & \textbf{36.0} & 20.2 & 27.3 & 50.9 & 46.4 & 28.4 & 6.7  & 37.2 \\
Ours + BoxPC          & 28.4 & 67.9 & 73.3 & 32.3 & 23.3 & 31.0 & 50.9 & 48.9 & 33.7 & \textbf{16.4} & 40.6 \\
Ours + BoxPC + R      & 28.4 & 68.1 & \textbf{77.9} & 32.9 & 23.3 & 30.6 & 51.1 & \textbf{49.9} & \textbf{34.7} & 13.7 & 41.1 \\
Ours + BoxPC + R + P  & \textbf{30.2} & \textbf{70.7} & 76.3 & 33.6 & \textbf{24.0} & \textbf{32.5} & \textbf{52.0} & 49.8 & 34.2 & 14.8 & \textbf{41.8} \\
\hline
\end{tabular}
\end{center}
   \caption{3D object detection AP on SUN-RGBD \textit{val} set. Fully-supervised methods are trained on 2D and 3D box labels of all classes while Cross-category (CC) Semi-supervised methods are trained on 3D box labels of classes in $C_{3D}$ and 2D box labels of all classes. BoxPC, R, P refer to the BoxPC Fit network, relaxed reprojection loss and box prior loss respectively. * refers to our implementation. }
\label{tab:sun_result_1}
\vspace{-0.1in}
\end{table*}

% \begin{table*}
% \begin{center}
% \begin{tabular}{l|c c c c c c c c c c|c}
% \hline
%  & bathtub & bed & bkshelf & chair & desk & dresser & nstand & sofa & table & toilet & mAP \\
% \hline
% \textbf{Fully-Supervised:} &  &  &  &  &  &  &  &  &  &  & \\
% FPN*         &  &  &  &  &  &  &  &  &  &  & \\
% \hline
% \textbf{CC Semi-Supervised:} &  &  &  &  &  &  &  &  &  &  & \\
% FPN*          &  &  &  &  &  &  &  &  &  &  & \\
% FPN* - OneHot &  &  &  &  &  &  &  &  &  &  & \\
% Ours + R              &  &  &  &  &  &  &  &  &  &  & \\
% Ours + BoxPC          &  &  &  &  &  &  &  &  &  &  & \\
% Ours + BoxPC + R      &  &  &  &  &  &  &  &  &  &  & \\
% Ours + BoxPC + R + IS &  &  &  &  &  &  &  &  &  &  & \\
% \hline
% \end{tabular}
% \end{center}
%   \caption{3D object detection AP on SUN-RGBD val set when given the ground truth 2D object bounding boxes during inference, this shows the potential improvement in 3D object detection performance with an improvement in the 2D object detector. }
% \label{tab:sun_result_2}
% \end{table*}

\vspace{-0.05in}
\subsubsection{Relaxed Reprojection Loss}
Despite the correlation, the 2D box that encloses all the points of a projected 3D box label does not coincide with the 2D box label. 
We show in Sec.~\ref{exp_ablation} that performance deteriorates if we simply minimize the reprojection loss between the 2D box label and the 2D box enclosing the projection of the predicted 3D box onto the image plane.
Instead, we propose a relaxed reprojection loss, where boundaries close to the 2D box labels are not penalized.
Let $B^*_{2D} = [b^*_{left}, b^*_{top}, b^*_{right}, b^*_{bot}]$ (green box in Fig.~\ref{fig:reproj_loss}) that consists of the left, top, right and bottom image coordinates be the 2D box label with the top-left hand corner as origin.
We define an upper bound box $B^*_{upper} = B^*_{2D} + \mathcal{U}$ and a lower bound box $B^*_{lower} = B^*_{2D} + \mathcal{L}$ (outer and inner blue boxes in Fig. \ref{fig:reproj_loss}), where $\mathcal{U} = [u_1, u_2, u_3, u_4]$ and $\mathcal{L} = [v_1, v_2, v_3, v_4]$ are vectors used to adjust the size of $B^*_{2D}$. Given an initial 3D box prediction $\hat{B}_0$, we project it onto the image plane $\pi(\hat{B}_0)$ (red box in Fig. \ref{fig:reproj_loss}) and obtain an enclosing 2D box around the projected points $\hat{B}_0^{reproj} = g(\pi(\hat{B}_0))$ (yellow box in Fig. \ref{fig:reproj_loss}), where $g(.)$ is the function that returns the enclosing 2D box. We penalize $\hat{B}_0^{reproj}$ for violating the bounds specified by $B^*_{upper}$ and $B^*_{lower}$. An example is shown in Fig.~\ref{fig:reproj_loss}, the bounds on the left and right are not violated because the left and right sides of the 2D box $\hat{B}_0^{reproj}$ stays within the bounds specified by $B^*_{upper}$ and $B^*_{lower}$ (blue boxes). However, the top and bottom of $\hat{B}_0^{reproj}$ beyond the bounds are penalized. More formally, the relaxed reprojection loss is given by
\vspace{-0.05in}
\begin{align}
  L_{reproj}^{2D} &= \sum_{c \in C_{2D}} \sum_{i = 1}^4 \tilde{l}(v_i,\, u_i,\, \hat{B}_0^{c,reproj}(i)), \label{eq:Loss_reproj2d} \\
  \tilde{l}(y^*_{lower}&, y^*_{upper}, \hat{y}) = 
    \begin{cases}
    l(y^*_{upper}, \hat{y}), &\text{if $ \hat{y} > y^*_{upper} $},\\
    l(y^*_{lower}, \hat{y}), &\text{if $ \hat{y} < y^*_{lower} $},\\
    0           &\text{otherwise.}
    \end{cases}
\end{align}

\vspace{-0.1in}
\hspace{-0.2in} $\tilde{l}(y^*_{lower}, y^*_{upper}, \hat{y})$ is a relaxed Smooth L1 loss such that there is no penalty on $\hat{y}$ if $y^*_{lower} \leq \hat{y} \leq y^*_{upper}$, and $\hat{B}_0^{c,reproj}(i)$ retrieves the $i$-th component of $\hat{B}_0^{c,reproj}$.

\vspace{-0.1in}
\subsubsection{Box Prior Loss}
We use prior knowledge on the object volume and size to regularize the training loss. More specifically, we add the prior loss 
\vspace{-0.1in}
\begin{equation}
L_{prior}^{2D} = \sum_{c \in C_{2D}} L_{vol}^c + L_{s-var}^c,    
\label{eq:Loss_prior2D}
\vspace{-0.1in}
\end{equation}
\vspace{-0.05in}
to train our network, where 
\begin{equation}
    L_{vol}^c = \max(0, V^c - \hat{b}_{0,h}^{\, c}\; \hat{b}_{0,w}^{\, c}\; \hat{b}_{0,l}^{\, c}),
\end{equation}
is to penalize predictions with volumes that are below class-specific thresholds
from prior knowledge about the scale of objects, and 
\vspace{-0.05in}
\begin{equation}
      L_{s-var}^c = l(\bar{b}_{0,h}^{\, c}, \hat{b}_{0,h}^{\, c}) + l(\bar{b}_{0,w}^{\, c}, \hat{b}_{0,w}^{\, c}) + l(\bar{b}_{0,l}^{\, c}, \hat{b}_{0,l}^{\, c}),
\end{equation}
is to penalize the size variance of the predictions within each class in each minibatch since there should not be excessive size variations within a class. $(\hat{b}_{0,h}^{\, c}, \hat{b}_{0,w}^{\, c}, \hat{b}_{0,l}^{\, c})$ are the size predictions from $\hat{B}_0$ for class $c$, and $(\bar{b}_{0,h}^{\, c}, \bar{b}_{0,w}^{\, c}, \bar{b}_{0,l}^{\, c})$ are the average size predictions per minibatch for class $c$ and $V^c$ is the volume threshold for class $c$.

%-------------------------------------------------------------------------
% Discuss loss functions over here
\section{Training}
We train a Faster RCNN-based \cite{FRCNN} 2D object detector with the 2D box labels from classes $C$ and train the BoxPC Fit Network by minimizing $L_{boxpc}^{3D}$ from Eq. \ref{eq:L_boxpc3d} with the 3D box labels from classes $C_{3D}$.
Finally, when we train the 3D object detector, we alternate between optimizing the losses for classes in $C_{2D}$ and classes in $C_{3D}$. Specifically, when optimizing for $C_{2D}$, we train $f_{box}$ to minimize 
\begin{equation}
    L_{box}^{2D} = L_{fit}^{2D} + L_{reproj}^{2D} + L_{prior}^{2D},
\end{equation}
where $L_{fit}^{2D}$, $L_{reproj}^{2D}$ and $L_{prior}^{2D}$ are from Eq. \ref{eq:Loss_fit2d}, \ref{eq:Loss_reproj2d} and \ref{eq:Loss_prior2D} respectively. 
When optimizing for $C_{3D}$, we train $f_{seg}$ and $f_{box}$ to minimize $L_{seg}^{3D}$ and $L_{box}^{3D}$ from Eq. \ref{eq:Loss_s} and \ref{eq:L_box3d} respectively.
Hence, the loss functions for $f_{seg}$ and $f_{box}$ are respectively given by
\begin{align}
L_{seg} &= L_{seg}^{3D}, \\
L_{box} &= L_{box}^{3D} + L_{box}^{2D}.
\end{align}

%-------------------------------------------------------------------------
\section {Experiments}

\begin{table*}
\begin{center}
\begin{tabular}{l||c c c||c c c||c c c}
\hline
\multirow{2}{*}{Method} & \multicolumn{3}{c||}{Cars} & \multicolumn{3}{c||}{Pedestrians} & \multicolumn{3}{c}{Cyclists}\\
\cline{2-10}
 & Easy & Moderate & Hard & Easy & Moderate & Hard & Easy & Moderate & Hard \\
\hline
\textbf{Fully-Supervised:} &  &  &  &  &  &  &  &  &  \\
FPN \cite{FPN}    & 96.69 & 96.29 & 88.77 & 85.40 & 79.20 & 71.61 & 86.59 & 67.60 & 63.95 \\
\hline
\textbf{CC Semi-Supervised:} &  &  &  &  &  &  &  &  & \\
FPN                  & 0.00  & 0.00  & 0.00 & 4.31  & 4.37  & 3.74   & 1.05  & 0.71  & 0.71 \\
FPN w/o OneHot       & 0.08  & 0.07  & 0.07 & 55.66 & 47.33 & 41.10 & 54.35 & 36.46 & 34.85 \\
Ours + BoxPC + R     & 14.71 & 12.19 & 11.21 & 75.13 & 62.46 & 57.18 & 61.25 & 41.96 & 40.07 \\
Ours + BoxPC + R + P & \textbf{69.78} & \textbf{58.66} & \textbf{51.40} & \textbf{76.85} & \textbf{66.92} & \textbf{58.71} & \textbf{63.29} & \textbf{47.54} & \textbf{44.66} \\
\hline
\end{tabular}
\end{center}
   \caption{3D object detection AP on KITTI \textit{val} set, evaluated at 3D IoU threshold of 0.25 for all classes. }
\label{tab:kitti_result_1}

\begin{center}
\begin{tabular}{l||c c c||c c c||c c c}
\hline
\multirow{2}{*}{Method} & \multicolumn{3}{c||}{Cars} & \multicolumn{3}{c||}{Pedestrians} & \multicolumn{3}{c}{Cyclists}\\
\cline{2-10}
 & Easy & Moderate & Hard & Easy & Moderate & Hard & Easy & Moderate & Hard \\
\hline
\textbf{Fully-Supervised:} &  &  &  &  &  &  &  &  &  \\
% 3DOP \cite{3DOP}         & 12.63 & 9.49  & 7.59  & - & - & - & - & - & - \\
VeloFCN \cite{VeloFCN}   & 15.20 & 13.66 & 15.98 & - & - & - & - & - & - \\
MV3D \cite{MV3D}         & 71.29 & 62.68 & 56.56 & - & - & - & - & - & - \\
Voxelnet \cite{VoxelNet} & \textbf{89.60} & \textbf{84.81} & \textbf{78.57} & 65.95 & \textbf{61.05} & \textbf{56.98} & 74.41 & 52.18 & 50.49 \\
FPN \cite{FPN}         & 84.40 & 71.37 & 63.38 & 65.80 & 56.40 & 49.74 & 76.11 & 56.88 & 53.17 \\
FPN     + BoxPC Refine & 85.63 & 72.10 & 64.25 & \textbf{68.83} & 59.41 & 52.08 & \textbf{78.77} & \textbf{58.41} & \textbf{54.52} \\
\hline
\end{tabular}
\end{center}
   \caption{3D object detection AP on KITTI \textit{val} set, evaluated at 3D IoU threshold of 0.7 for \textit{Cars} and 0.5 for \textit{Pedestrians} and \textit{Cyclists}. }
\label{tab:kitti_result_2}
\vspace{-0.1in}
\end{table*}

\subsection{Datasets} \label{exp_datasets}
We evaluate our model on SUN-RGBD and KITTI benchmarks for 3D object detection. For the SUN-RGBD benchmark, we evaluate on 10 classes as in \cite{FPN}. To test the performance of cross-category semi-supervised 3D object detection (CS3D), we randomly split the 10 classes into 2 subsets ($C_A$ and $C_B$) of 5 classes each (the 5 classes on the left and on the right of Tab.~\ref{tab:sun_result_1}). 
First, we let $C_A$ be the \textit{strong} classes and $C_B$ be the \textit{weak} classes by setting $C_{3D} = C_A, C_{2D} = C_{B}$ to obtain the evaluation results for $C_B$. Next, we reversed $C_A$ and $C_B$ to obtain the evaluation results for $C_A$.
We follow the same \textit{train}/\textit{val} split and performance measure of average precision (AP) with 3D Intersection-over-Union (IoU) of 0.25 as in \cite{FPN}.

The KITTI benchmark evaluates on \textit{Cars}, \textit{Pedestrians} and \textit{Cyclists} classes. To test the CS3D setting, we set any 2 classes to $C_{3D}$ and the remaining class to $C_{2D}$ to obtain the evaluation results on $C_{2D}$. For example, to obtain the evaluation results on \textit{Cyclists}, we set $C_{3D} = \text{ \{\textit{Cars}, \textit{Pedestrians}\} } $. We follow the same \textit{train}/\textit{val} split as in \cite{FPN} and measure performance using AP with 3D IoU of 0.25 for all classes. For fully-supervised setting, we use 3D IoU of 0.7 for \textit{Cars} and 0.5 for \textit{Pedestrians} and \textit{Cyclists}.

\subsection{Comparison with Baselines}

To the best of our knowledge, there are no other methods that demonstrate weakly- or semi-supervised 3D object detection, therefore we design baselines (details in supplementary) with the state-of-the-art FPN \cite{FPN} where it trains on strong labels from $C_{3D}$ and tests on $C_{2D}$.
Specifically, we use the original network (``FPN*'') and a network without the one hot class vector (``FPN* w/o OneHot''). The latter performs better since there are no 3D box labels for classes $C_{2D}$.
Finally, we train fully-supervised BoxPC Fit networks to further improve the fully-supervised FPN \cite{FPN}.

\vspace{-0.125in}
\paragraph{SUN-RGBD} \label{exp_sunrgbd} In Tab.~\ref{tab:sun_result_1}, BoxPC, R, P refer to the usage of the proposed BoxPC Fit network, relaxed reprojection loss and box prior loss. 
{\color{black} The fully-supervised ``FPN*'' serves as an upper-bound performance for the semi-supervised methods. }
The semi-supervised baselines ``FPN*'' and ``FPN* w/o OneHot'' perform poorly because they are not able to reason about the \textit{weak} classes and predict boxes that are only roughly correct. 
By adding R and BoxPC, it allows the network to reason about 3D box predictions on \textit{weak} classes, improving performance significantly from 30.1\% to 41.1\%. 
{\color{black} Finally, the addition of the prior knowledge P allows our model to achieve a good performance of 41.8\% mAP, which is 81.2\% of the fully-supervised ``FPN*'' (vs 58.4\% for baseline ``FPN* w/o OneHot''). We consider this a promising result for semi-supervised methods, and it demonstrates the effectiveness of our proposed model in transferring knowledge to unseen classes.}
% Finally, the addition of our prior knowledge P allows the proposed model to achieve a strong performance of 41.8\% that is comparable to many fully-supervised methods such as \cite{DeepSlidingShapes,COG,2DDriven3D}, demonstrating the effectiveness of our proposed model in transferring knowledge across classes.
In addition, we show the effectiveness of the BoxPC Fit network in fully-supervised settings by training the BoxPC Fit network on all classes $C$ and using the network to refine the 3D box predictions of the state-of-the-art FPN \cite{FPN}. 
As seen in Tab.~\ref{tab:sun_result_1}, ``FPN* + BoxPC Refine'' outperforms the vanilla fully-supervised ``FPN*'' in  every single class except for \texttt{bookshelf}, showing the usefulness of the BoxPC Fit network even in fully-supervised settings.

\vspace{-0.125in}
\paragraph{KITTI} \label{exp_kitti} In Tab.~\ref{tab:kitti_result_1}, the baselines are unable to achieve any performance on \textit{Cars} when trained on \textit{Pedestrians} and \textit{Cyclists} because of huge differences in sizes. The network simply assumes a \textit{Car} instance is small and makes small predictions. With the addition of a prior volume, it becomes possible to make predictions on \textit{Cars}. We observe that adding ``BoxPC + R + P'' significantly improves performance over the baseline's mean AP of 0.07\% to 59.95\% for \textit{Cars}. It also improved the baseline's performance from 48.03\% to 67.49\% for \textit{Pedestrians} and from 41.89\% to 51.83\% for \textit{Cyclists}.
Similarly, we show that a fully-supervised BoxPC Fit network is able to refine and improve the 3D box predictions made by ``FPN'' in Tab.~\ref{tab:kitti_result_2}. In this case, ``FPN + BoxPC Refine'' improves AP for all classes at all difficulty levels over the original ``FPN''.

\subsection{Ablation Studies} \label{exp_ablation}

\begin{table*}
\begin{center}
\begin{tabular}{c|c|c c c c c c c c c c|c}
\hline
$B^*_{lower}$ & $B^*_{upper}$ & bathtub & bed & toilet & chair & desk & dresser & nstand & sofa & table & bkshelf & mAP \\
\hline
$B^*_{2D}$ & $1.0\ B^*_{2D}$   & 17.0 & 11.9 & 60.1 & 24.3 & 11.2 & 24.0 & 46.8 & 30.3 & 19.2 & \textbf{11.3} & 25.6 \\
$B^*_{2D}$ & $1.5\ B^*_{2D}$ & \textbf{29.5} & \textbf{60.9} & \textbf{65.3} & \textbf{36.0} & \textbf{20.2} & \textbf{27.3} & \textbf{50.9} & 46.4 & 28.4 & 6.7  & \textbf{37.2} \\
$B^*_{2D}$ & $2.0\ B^*_{2D}$ & 16.9 & 23.4 & 55.8 & 35.0 & 19.6 & 25.1 & 49.2 & \textbf{48.3} & \textbf{31.3} & 3.5  & 30.8 \\
\hline
\end{tabular}
\end{center}
   \caption{3D object detection AP on SUN-RGBD \textit{val} set with varying $B^*_{upper}$ for the relaxed reprojection loss function.}
\label{tab:sun_abla_reproj}

\begin{center}
\begin{tabular}{c|c|c|c c c c c c c c c c|c}
\hline
Features & Cls & Reg & bathtub & bed & toilet & chair & desk & dresser & nstand & sofa & table & bkshelf & mAP \\
\hline
Combined    & \checkmark &            & 25.9 & 67.3 & 71.8 & 30.6 & \textbf{23.6} & 31.8 & 48.8 & 48.1 & 32.6 & \textbf{17.3} & 39.8 \\
Combined    &            & \checkmark & 27.4 & \textbf{69.4} & 39.8 & 17.7 & 23.1 & 18.5 & 39.9 & 41.7 & 30.2 & 12.9 & 32.1 \\
Independent & \checkmark & \checkmark & \textbf{39.6} & 68.1 & 24.5 & 26.8 & 22.3 & \textbf{34.4} & 38.8 & 46.7 & 30.3 & 12.3 & 34.4 \\
Combined    & \checkmark & \checkmark & 28.4 & 67.9 & \textbf{73.3} & \textbf{32.3} & 23.3 & 31.0 & \textbf{50.9} & \textbf{48.9} & \textbf{33.7} & 16.4 & \textbf{40.6} \\
\hline
\end{tabular}
\end{center}
   \caption{3D object detection AP on SUN-RGBD \textit{val} set when BoxPC Fit Network is trained on different representations and objectives. }
\label{tab:sun_abla_boxpc}
\end{table*}

\begin{figure*}
\begin{center}
\includegraphics[width=\linewidth]{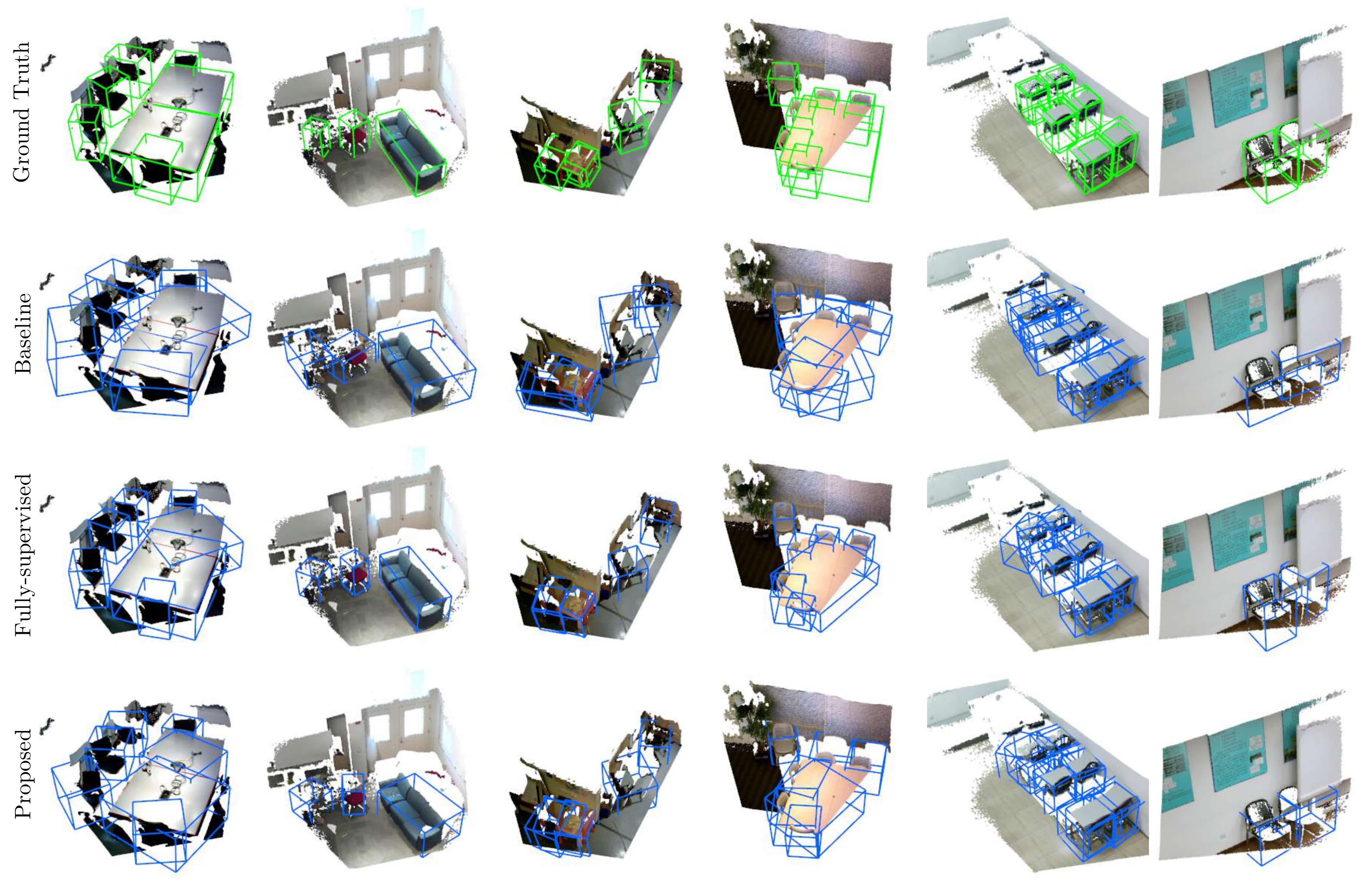}
\end{center}
   \caption{Qualitative comparisons between the baseline, fully-supervised and proposed models on the SUN-RGBD \textit{val} set.}
\label{fig:exp_qualitative}
\end{figure*} 

\vspace{-0.05in}
\paragraph{Relaxed Reprojection Loss} To illustrate the usefulness of using a relaxed version of the reprojection loss, we vary the lower and upper bounds $B^*_{lower}$ and $B^*_{upper}$. We set $B^*_{lower}, B^*_{upper}$ to different scales of $B^*_{2D}$, i.e $1.5\ B^*_{2D}$ refers to the 2D box centered at the same point as $B^*_{2D}$ but with 1.5$\times$ size. By setting $B^*_{lower}, B^*_{upper} = B^*_{2D}$, it reverts from the relaxed version to the direct version of reprojection loss. We train with similar settings as ``Ours + R'' in Tab.~\ref{tab:sun_result_1}, and show in Tab.~\ref{tab:sun_abla_reproj} that mean AP improves from 25.6\% ($B^*_{upper} = B^*_{2D}$) to 37.2\% ($B^*_{upper} = 1.5\ B^*_{2D}$).

\vspace{-0.1in}
\paragraph{BoxPC Fit Network Objectives} The BoxPC Fit network is trained on classification and regression objectives derived from the 3D box labels. To show the effectiveness of having both objectives, we adopt different settings in which classification or regression objectives are active. We train with similar settings as ``Ours + BoxPC'' in Tab.~\ref{tab:sun_result_1}. Without the classification objective for the BoxPC Fit network, the 3D object detector is unable to maximize $\hat{p} = f_{boxpc-cls}(X, \hat{B}_0)$ during training. 
Omission of the regression objective prevents the initial box predictions $\hat{B}_0$ from being refined further by $\Delta \hat{B} = f_{boxpc-reg}(X, \hat{B}_0)$ to the final box predictions $\hat{B}$. 
From Tab.~\ref{tab:sun_abla_boxpc}, we see that training without either classification or regression objective causes performance to drop from 40.6\% to 32.1\% or 39.8\%.

\vspace{-0.1in}
\paragraph{BoxPC Representation} To demonstrate the importance of a joint Box to Point Cloud input representation as seen on the right of Fig.~\ref{fig:network_boxpc}, we train on the ``Ours + BoxPC'' setting in Tab.~\ref{tab:sun_result_1} with different BoxPC representations. As shown in Tab.~\ref{tab:sun_abla_boxpc}, the performance of combined learning of BoxPC features exceeds the performance of independent learning of BoxPC features by 6.2\%.

\subsection{Qualitative Results} \label{exp_qual}
We visualize some of the predictions by the baseline, fully-supervised and proposed models in Fig.~\ref{fig:exp_qualitative}. We used 2D box labels for frustum proposals to reduce noise due to 2D detections. We observe that the proposed model's predictions are closer to the fully-supervised model than the baseline's. The baseline's predictions tend to be inaccurate since it is not trained to fit the objects from \textit{weak} classes.

%-------------------------------------------------------------------------
\section {Conclusion}
We propose a transferable semi-supervised model that is able to perform 3D object detection on \textit{weak} classes with only 2D box labels. We achieve strong performance over the baseline approaches and in the fully-supervised setting, we improved the performance of existing detectors. In conclusion, our method improves the practicality and usefulness of 3D object detectors in new applications.

{\small
\bibliographystyle{ieee}
\bibliography{egbib}

\begin{thebibliography}{10}\itemsep=-1pt

\bibitem{NetVLAD}
R.~Arandjelovic, P.~Gronat, A.~Torii, T.~Pajdla, and J.~Sivic.
\newblock Netvlad: Cnn architecture for weakly supervised place recognition.
\newblock In {\em Proceedings of the IEEE Conference on Computer Vision and
  Pattern Recognition}, pages 5297--5307, 2016.

\bibitem{Weakly3DHPE}
Y.~Cai, L.~Ge, J.~Cai, and J.~Yuan.
\newblock Weakly-supervised 3d hand pose estimation from monocular rgb images.
\newblock In {\em European Conference on Computer Vision (ECCV)}, 2018.

\bibitem{BeatMTurkers}
L.-C. Chen, S.~Fidler, A.~L. Yuille, and R.~Urtasun.
\newblock Beat the mturkers: Automatic image labeling from weak 3d supervision.
\newblock In {\em Proceedings of the IEEE Conference on Computer Vision and
  Pattern Recognition}, pages 3198--3205, 2014.

\bibitem{FRCNN_github}
X.~Chen and A.~Gupta.
\newblock An implementation of faster rcnn with study for region sampling.
\newblock {\em arXiv preprint arXiv:1702.02138}, 2017.

\bibitem{Mono3D}
X.~Chen, K.~Kundu, Z.~Zhang, H.~Ma, S.~Fidler, and R.~Urtasun.
\newblock Monocular 3d object detection for autonomous driving.
\newblock In {\em Proceedings of the IEEE Conference on Computer Vision and
  Pattern Recognition}, pages 2147--2156, 2016.

\bibitem{3DOP}
X.~Chen, K.~Kundu, Y.~Zhu, A.~G. Berneshawi, H.~Ma, S.~Fidler, and R.~Urtasun.
\newblock 3d object proposals for accurate object class detection.
\newblock In {\em Advances in Neural Information Processing Systems}, pages
  424--432, 2015.

\bibitem{MV3D}
X.~Chen, H.~Ma, J.~Wan, B.~Li, and T.~Xia.
\newblock Multi-view 3d object detection network for autonomous driving.
\newblock In {\em IEEE CVPR}, volume~1, page~3, 2017.

\bibitem{Kitti}
A.~Geiger, P.~Lenz, and R.~Urtasun.
\newblock Are we ready for autonomous driving? the kitti vision benchmark
  suite.
\newblock In {\em Computer Vision and Pattern Recognition (CVPR), 2012 IEEE
  Conference on}, pages 3354--3361. IEEE, 2012.

\bibitem{Decoupled}
S.~Hong, H.~Noh, and B.~Han.
\newblock Decoupled deep neural network for semi-supervised semantic
  segmentation.
\newblock In {\em Advances in neural information processing systems}, pages
  1495--1503, 2015.

\bibitem{Trans2}
S.~Hong, J.~Oh, H.~Lee, and B.~Han.
\newblock Learning transferrable knowledge for semantic segmentation with deep
  convolutional neural network.
\newblock In {\em Proceedings of the IEEE Conference on Computer Vision and
  Pattern Recognition}, pages 3204--3212, 2016.

\bibitem{Trans3}
R.~Hu, P.~Doll{\'a}r, K.~He, T.~Darrell, and R.~Girshick.
\newblock Learning to segment every thing.
\newblock In {\em Proceedings of the IEEE Conference on Computer Vision and
  Pattern Recognition}, pages 4233--4241, 2018.

\bibitem{SEC}
A.~Kolesnikov and C.~H. Lampert.
\newblock Seed, expand and constrain: Three principles for weakly-supervised
  image segmentation.
\newblock In {\em European Conference on Computer Vision (ECCV)}, pages
  695--711. Springer, 2016.

\bibitem{AVOD}
J.~Ku, M.~Mozifian, J.~Lee, A.~Harakeh, and S.~Waslander.
\newblock Joint 3d proposal generation and object detection from view
  aggregation.
\newblock {\em arXiv preprint arXiv:1712.02294}, 2017.

\bibitem{2DDriven3D}
J.~Lahoud and B.~Ghanem.
\newblock 2d-driven 3d object detection in rgb-d images.
\newblock In {\em 2017 IEEE International Conference on Computer Vision
  (ICCV)}, pages 4632--4640. IEEE, 2017.

\bibitem{VeloFCN}
B.~Li.
\newblock 3d fully convolutional network for vehicle detection in point cloud.
\newblock In {\em Intelligent Robots and Systems (IROS), 2017 IEEE/RSJ
  International Conference on}, pages 1513--1518. IEEE, 2017.

\bibitem{WeakSemiPanop}
Q.~Li, A.~Arnab, and P.~H. Torr.
\newblock Weakly- and semi-supervised panoptic segmentation.
\newblock In {\em European Conference on Computer Vision (ECCV)}, 2018.

\bibitem{Deep3DBox}
A.~Mousavian, D.~Anguelov, J.~Flynn, and J.~Ko{\v{s}}eck{\'a}.
\newblock 3d bounding box estimation using deep learning and geometry.
\newblock In {\em Computer Vision and Pattern Recognition (CVPR), 2017 IEEE
  Conference on}, pages 5632--5640. IEEE, 2017.

\bibitem{ExtremeClicking}
D.~P. Papadopoulos, J.~R. Uijlings, F.~Keller, and V.~Ferrari.
\newblock Extreme clicking for efficient object annotation.
\newblock In {\em Proceedings of the ICCV}, pages 4940--4949. IEEE, 2017.

\bibitem{WeakAndSemiSemSeg}
G.~Papandreou, L.-C. Chen, K.~P. Murphy, and A.~L. Yuille.
\newblock Weakly-and semi-supervised learning of a deep convolutional network
  for semantic image segmentation.
\newblock In {\em Proceedings of the IEEE international conference on computer
  vision}, pages 1742--1750, 2015.

\bibitem{CCNN}
D.~Pathak, P.~Krahenbuhl, and T.~Darrell.
\newblock Constrained convolutional neural networks for weakly supervised
  segmentation.
\newblock In {\em Proceedings of the IEEE international conference on computer
  vision}, pages 1796--1804, 2015.

\bibitem{FPN}
C.~R. Qi, W.~Liu, C.~Wu, H.~Su, and L.~J. Guibas.
\newblock Frustum pointnets for 3d object detection from rgb-d data.
\newblock In {\em Proceedings of the IEEE Conference on Computer Vision and
  Pattern Recognition}, pages 918--927, 2018.

\bibitem{PN}
C.~R. Qi, H.~Su, K.~Mo, and L.~J. Guibas.
\newblock Pointnet: Deep learning on point sets for 3d classification and
  segmentation.
\newblock {\em Proc. Computer Vision and Pattern Recognition (CVPR), IEEE},
  1(2):4, 2017.

\bibitem{FRCNN}
S.~Ren, K.~He, R.~Girshick, and J.~Sun.
\newblock Faster r-cnn: Towards real-time object detection with region proposal
  networks.
\newblock In {\em Advances in neural information processing systems}, pages
  91--99, 2015.

\bibitem{COG}
Z.~Ren and E.~B. Sudderth.
\newblock Three-dimensional object detection and layout prediction using clouds
  of oriented gradients.
\newblock In {\em Proceedings of the IEEE Conference on Computer Vision and
  Pattern Recognition}, pages 1525--1533, 2016.

\bibitem{LSS}
Z.~Ren and E.~B. Sudderth.
\newblock 3d object detection with latent support surfaces.
\newblock In {\em Proceedings of the IEEE Conference on Computer Vision and
  Pattern Recognition}, pages 937--946, 2018.

\bibitem{SUN}
S.~Song, S.~P. Lichtenberg, and J.~Xiao.
\newblock Sun rgb-d: A rgb-d scene understanding benchmark suite.
\newblock In {\em Proceedings of the IEEE conference on computer vision and
  pattern recognition}, pages 567--576, 2015.

\bibitem{DeepSlidingShapes}
S.~Song and J.~Xiao.
\newblock Deep sliding shapes for amodal 3d object detection in rgb-d images.
\newblock In {\em Proceedings of the IEEE Conference on Computer Vision and
  Pattern Recognition}, pages 808--816, 2016.

\bibitem{WeakShapeCompletionFromLaser}
D.~Stutz and A.~Geiger.
\newblock Learning 3d shape completion from laser scan data with weak
  supervision.
\newblock In {\em Proceedings of the IEEE Conference on Computer Vision and
  Pattern Recognition}, pages 1955--1964, 2018.

\bibitem{Implicit}
M.~Sundermeyer, Z.-C. Marton, M.~Durner, M.~Brucker, and R.~Triebel.
\newblock Implicit 3d orientation learning for 6d object detection from rgb
  images.
\newblock In {\em European Conference on Computer Vision (ECCV)}, pages
  712--729. Springer, 2018.

\bibitem{WeakRPN}
P.~Tang, X.~Wang, A.~Wang, Y.~Yan, W.~Liu, J.~Huang, and A.~Yuille.
\newblock Weakly supervised region proposal network and object detection.
\newblock In {\em European Conference on Computer Vision (ECCV)}, pages
  352--368, 2018.

\bibitem{RandomWalk}
P.~Vernaza and M.~Chandraker.
\newblock Learning random-walk label propagation for weakly-supervised semantic
  segmentation.
\newblock In {\em The IEEE Conference on Computer Vision and Pattern
  Recognition (CVPR)}, volume~3, 2017.

\bibitem{Vote3D}
D.~Z. Wang and I.~Posner.
\newblock Voting for voting in online point cloud object detection.
\newblock In {\em Robotics: Science and Systems}, volume~1, page~5, 2015.

\bibitem{ObjectNet3d}
Y.~Xiang, W.~Kim, W.~Chen, J.~Ji, C.~Choy, H.~Su, R.~Mottaghi, L.~Guibas, and
  S.~Savarese.
\newblock Objectnet3d: A large scale database for 3d object recognition.
\newblock In {\em European Conference Computer Vision (ECCV)}, 2016.

\bibitem{Pascal3D}
Y.~Xiang, R.~Mottaghi, and S.~Savarese.
\newblock Beyond pascal: A benchmark for 3d object detection in the wild.
\newblock In {\em Applications of Computer Vision (WACV), 2014 IEEE Winter
  Conference on}, pages 75--82. IEEE, 2014.

\bibitem{Trans1}
H.~Xiao, Y.~Wei, Y.~Liu, M.~Zhang, and J.~Feng.
\newblock Transferable semi-supervised semantic segmentation.
\newblock In {\em Association for the Advancement of Artificial Intelligence},
  2017.

\bibitem{PointFusion}
D.~Xu, D.~Anguelov, and A.~Jain.
\newblock Pointfusion: Deep sensor fusion for 3d bounding box estimation.
\newblock {\em arXiv preprint arXiv:1711.10871}, 2017.

\bibitem{LearningToSegUnderVariousWeak}
J.~Xu, A.~G. Schwing, and R.~Urtasun.
\newblock Learning to segment under various forms of weak supervision.
\newblock In {\em Computer Vision and Pattern Recognition (CVPR), 2015 IEEE
  Conference on}, pages 3781--3790. IEEE, 2015.

\bibitem{3DFeatNet}
Z.~J. Yew and G.~H. Lee.
\newblock 3dfeat-net: Weakly supervised local 3d features for point cloud
  registration.
\newblock In {\em European Conference on Computer Vision (ECCV)}, 2018.

\bibitem{3DShapeModelsFrom2DImages}
D.~Zhang, J.~Han, Y.~Yang, and D.~Huang.
\newblock Learning category-specific 3d shape models from weakly labeled 2d
  images.
\newblock In {\em Proc. CVPR}, pages 4573--4581, 2017.

\bibitem{3DHumanPoseWeak}
X.~Zhou, Q.~Huang, X.~Sun, X.~Xue, and Y.~Wei.
\newblock Towards 3d human pose estimation in the wild: a weakly-supervised
  approach.
\newblock In {\em IEEE International Conference on Computer Vision}, 2017.

\bibitem{VoxelNet}
Y.~Zhou and O.~Tuzel.
\newblock Voxelnet: End-to-end learning for point cloud based 3d object
  detection.
\newblock {\em arXiv preprint arXiv:1711.06396}, 2017.

\bibitem{WeakInstSegClassPeak}
Y.~Zhou, Y.~Zhu, Q.~Ye, Q.~Qiu, and J.~Jiao.
\newblock Weakly supervised instance segmentation using class peak response.
\newblock In {\em Proceedings of the IEEE international conference on computer
  vision}, pages 3791--3800, 2018.

\end{thebibliography}
}

\section*{Supplementary Materials}

\renewcommand{\thesubsection}{\Alph{subsection}}

{ \color{black} In Sec.~\ref{supp_costs}, we discuss the costs of different types of labels. In Sec.~\ref{supp_icss}, we observe the performance with varying amounts of 3D box labels for $C_{2D}$ classes. }
In Sec.~\ref{supp_network} and Sec.~\ref{supp_training}, we elaborate on the details of the networks and the details of the training procedures, respectively. In Sec.~\ref{supp_sun}, we provide additional qualitative results and figures on the SUN-RGBD dataset and in Sec.~\ref{supp_kitti}, qualitative results for the KITTI dataset.

{ \color{black} 
\subsection{Time Costs of Labels} \label{supp_costs}
The SUN-RGBD dataset \cite{SUN} required 2,051 hours to label 64,595 3D bboxes, which is an average of 114s per object.
In contrast, the average time to label a 2D bbox according to \cite{ExtremeClicking} is 35s per object but can be as fast as 7s when using their proposed labeling method with no loss of accuracy. 
%To put things in perspective, even object segmentation labels take only 79s \cite{ExtremeClicking} per object and is faster than 3D box labels. 
Hence, it is potentially 3-16 times faster to label 2D compared to 3D bboxes. 
%a 3D object detector that seeks to train on 2D box labels of $C_{2D}$ classes can potentially achieve 3-16 times savings in time as compared to using 3D box labels.
}

\begin{figure}%[hbt!]
\begin{center}
%   \fbox{\rule{0pt}{0.75in}} %\rule{0.9\linewidth}{0pt}}
  \includegraphics[width=\linewidth]{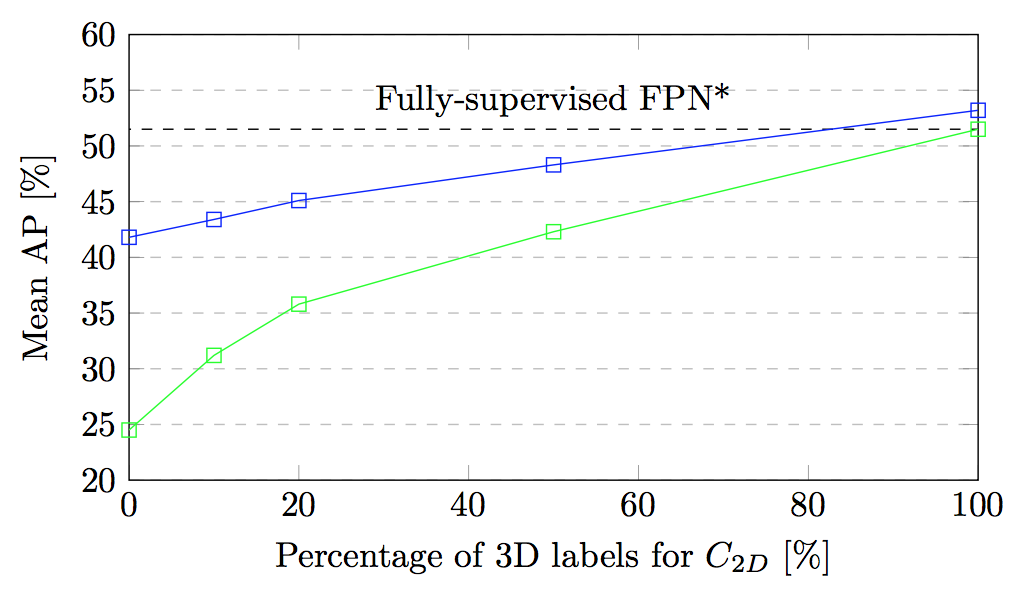}
  \vspace{-0.2in}
\end{center}
  \caption{{\color{black}3D object detection mAP on SUN-RGBD \textit{val} set of our proposed model with different percentages of 3D labels for $C_{2D}$. The blue and green lines correspond to the ``Ours + BoxPC + R + P'' and ``FPN*'' semi-supervised settings, respectively.}}
\label{fig:perf_with_diff_data}
\end{figure}

{ \color{black} 
\subsection{In-Category Semi-supervision Performance} \label{supp_icss}
In the main paper, we assumed that there are no 3D box labels for \textit{weak} classes $C_{2D}$, which is a cross-category semi-supervised setting.
In Fig.~\ref{fig:perf_with_diff_data}, we train our model and the baseline ``FPN*'' on varying amounts of 3D box labels for $C_{2D}$ to understand the performance of our model in an in-category semi-supervised setting. 
%where there are varying amounts of 3D box labels for the $C_{2D}$ classes. 
When the percentage of 3D box labels used for $C_{2D}$ is $100\%$, the semi-supervised baseline ``FPN*'' (green line) becomes the fully-supervised ``FPN*'' (51.5\% mAP) and our proposed semi-supervised method (blue line) becomes the fully-supervised ``FPN* + BoxPC Refine'' (53.2\% mAP).

We observe that our proposed method always performs better than the baseline for different percentages of 3D box labels of $C_{2D}$ classes. This demonstrates the usefulness of our method even when labels are available for $C_{2D}$ classes. Additionally, we note that the baseline with 50\% 3D box labels available for $C_{2D}$ achieves a similar performance to our proposed method with 0\% 3D box labels for $C_{2D}$, which demonstrates the effectiveness of the knowledge that has been transferred from $C_{3D}$ classes.
}

%\vspace{-0.2in}

\subsection{Network Details} \label{supp_network}

\subsubsection{Network Architecture of Baselines}
\begin{figure*}
\begin{center}
\includegraphics[width=\linewidth]{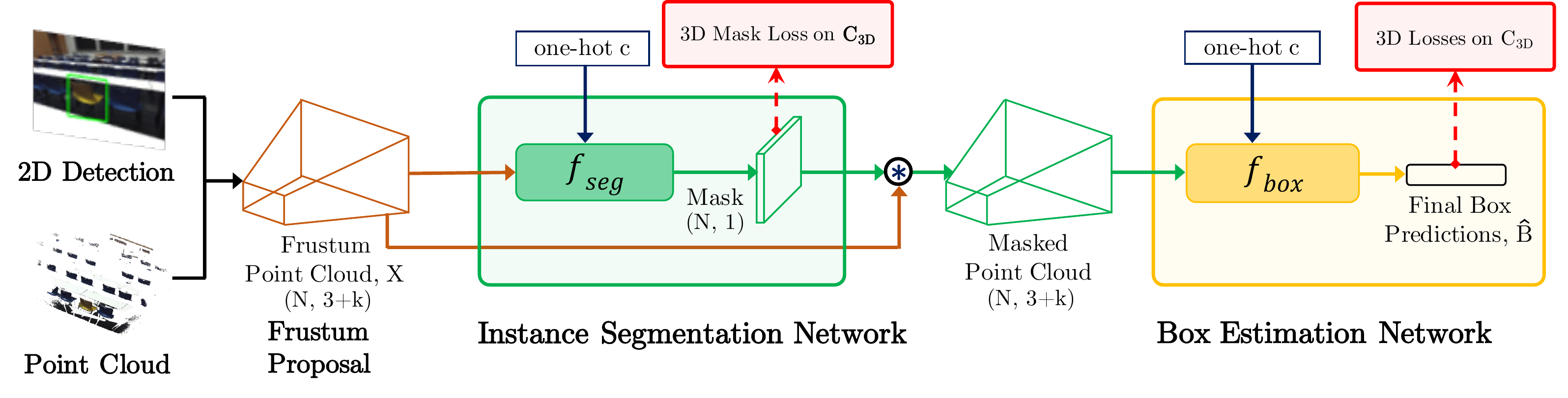}
\end{center}
   \caption{Network architecture for the original Frustum PointNets for fully-supervised 3D object detection. This network corresponds to ``FPN'' and the baseline ``FPN*'' in Tab.\ 1 of the main paper. To obtain the stronger baseline ``FPN* w/o OneHot'', we remove the one hot class vectors that are given as features to $f_{seg}$ and $f_{box}$.}
\label{fig:supp_network_baseline}
\vspace{0.1in}

\begin{center}
\includegraphics[width=\linewidth]{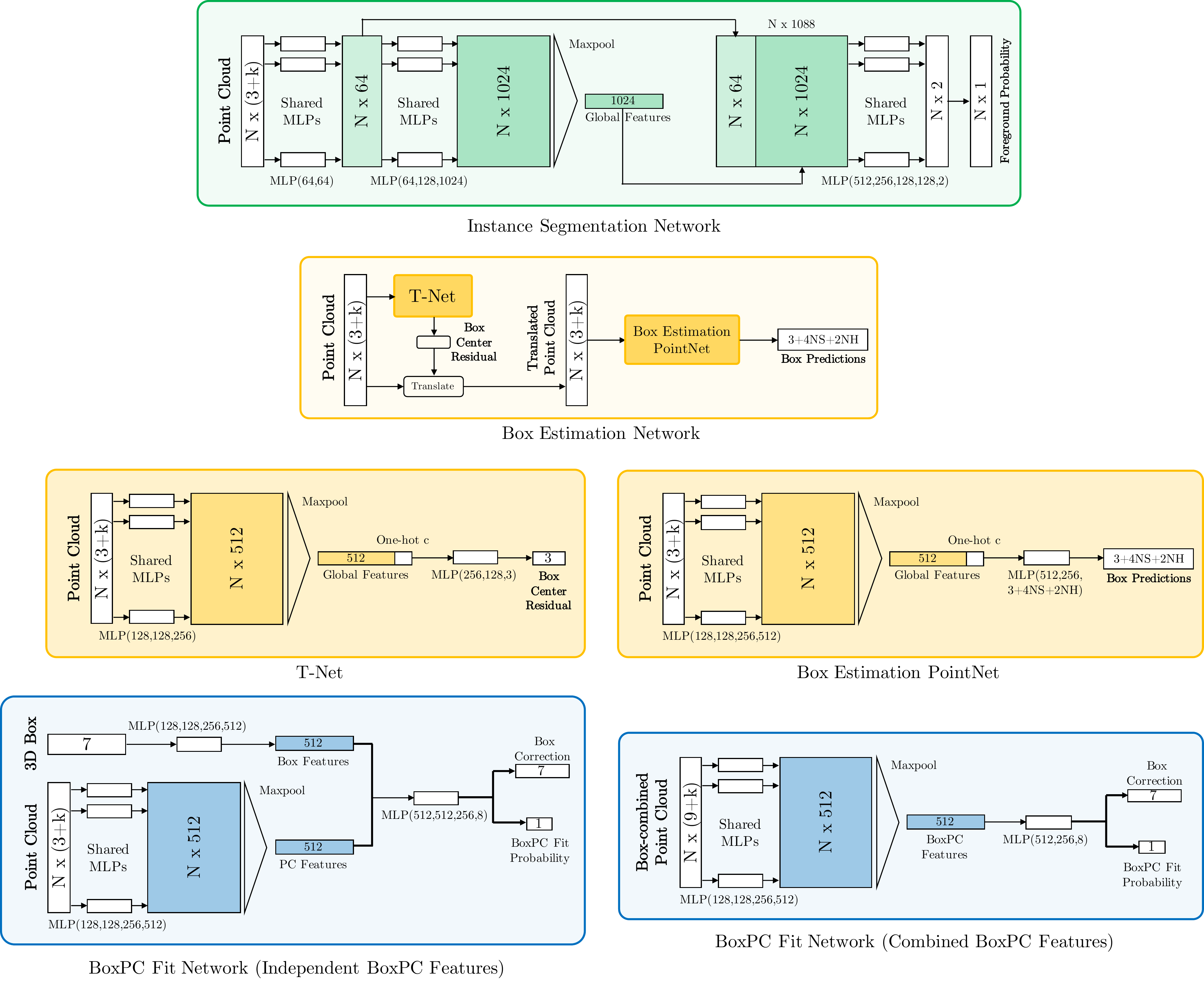}
\end{center}
   \caption{Network component details for our baseline and proposed models that are used in cross-category semi-supervised 3D object detection. In fully-supervised setting, a one hot class vector is added to the ``Global Features'' of the Instance Segmentation Network and the ``BoxPC Features'' of the BoxPC Fit Network. }
\label{fig:supp_network_components}
\end{figure*}

\vspace{-0.05in}
Fig.~\ref{fig:supp_network_baseline} shows the network architecture for the original Frustum PointNets (FPN) \cite{FPN}. The same architecture is used to obtain the results for ``FPN'' and the baseline ``FPN*'' in Tab. 1 of the main paper. We remove the one hot class vectors given as features to the $f_{seg}$ and $f_{box}$ networks to get the stronger baseline ``FPN* w/o OneHot''. The performance improves because in the cross-category semi-supervised setting, the network does not train on the strong labels of inference classes. Hence, having class information does not help the network during inference.

\begin{figure*}[t]
\begin{center}
   \includegraphics[width=\linewidth]{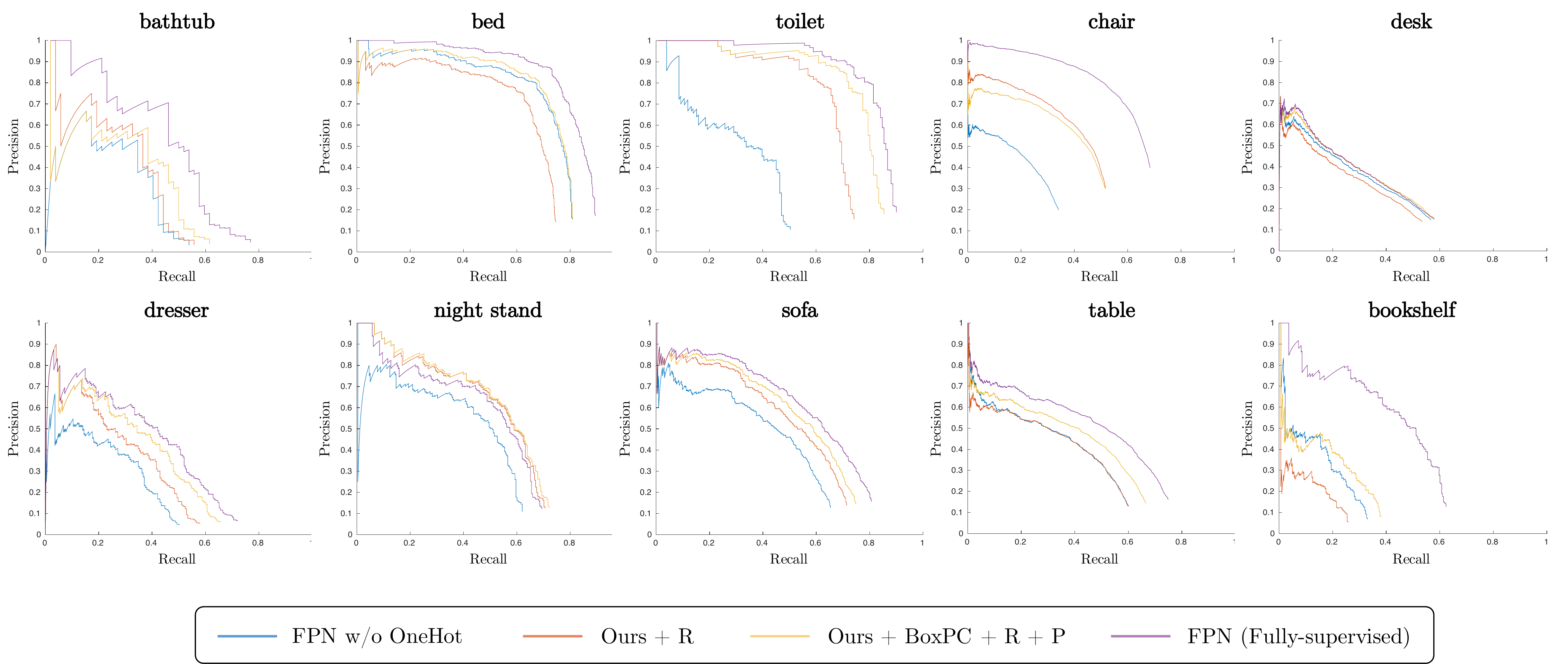}
\end{center}
   \caption{Precision recall (PR) curves for 3D object detection on SUN-RGBD \textit{val} set with different methods. ``FPN (Fully-supervised)'' is the fully-supervised model that gives an upper-bound performance for the rest of the models which are cross-category semi-supervised.}
\label{fig:supp_sun_pr_curves}
\vspace{-0.1in}
\end{figure*}

\vspace{-0.1in}
\subsubsection{Network Component Details}
\vspace{-0.05in}
In this section, we describe the details of the network components in the baseline and proposed models for the cross-category semi-supervised 3D object detection (CS3D) setting. The fully-supervised 3D object detection (FS3D) setting uses the same components except for minor changes. 

The network components are given in Fig.~\ref{fig:supp_network_components}. We use the v1 instance segmentation and v1 box estimation networks in FPN \cite{FPN} as the instance segmentation and box estimation networks in our baseline and proposed models. The BoxPC Fit networks also use PointNet \cite{PN} and MLP layers to learn the BoxPC features.

In the CS3D setting, we remove the one hot class vector that is originally concatenated with the ``Global Features'' of the v1 instance segmentation network because we are performing class-agnostic instance segmentations. In the FS3D setting, the one hot class vector is added back. 

The v1 box estimation network is composed of the T-Net and Box Estimation PointNet. The T-Net gives an initial prediction for the center of the box $b_x, b_y, b_z$ which is used to translate the input point cloud to reduce translational variance. Then, the Box Estimation PointNet makes the box predictions using the translated point cloud. Both CS3D and FS3D settings use the same box estimation networks. The details of the loss functions to train the box estimation networks can be found in \cite{FPN}.

In the CS3D setting, the BoxPC Fit network learns class-agnostic BoxPC Fit between 3D boxes and point clouds. In the FS3D setting, we concatenate a one hot class vector to the ``BoxPC Features'' to allow the network to learn BoxPC Fit that is specific to each class.

\subsection{Training Details} \label{supp_training}

\subsubsection{Training of BoxPC Fit Network}
\vspace{-0.05in}
As discussed in the paper, we have to sample from 2 sets of perturbations $\textbf{P}^+$ and $\textbf{P}^-$ to train the BoxPC Fit Network to understand what is a good BoxPC fit. To sample perturbations $\delta = [\delta_x, \delta_y, \delta_z, \delta_h, \delta_w, \delta_l, \delta_{\theta}]$ from either $\textbf{P}^+$ or $\textbf{P}^-$, we uniformly sample center perturbations $\delta_x, \delta_y, \delta_z \in [-0.8,0.8]$, size perturbations $\delta_h, \delta_w, \delta_l \in [-0.2,0.2]$ and rotation perturbations $\delta_{\theta} \in [0,\pi]$. We perturb a 3D box label $B^*$ to obtain a perturbed 3D box label $B^* - \delta$ with a sampled perturbation $\delta$.
Next, we check if $B^* - \delta$ has an IOU with $B^*$ that is within the range specified by the set $\textbf{P}^+$ or $\textbf{P}^-$, i.e if $ {\alpha}^+ \leq IOU(B^* - \delta, B^*) \leq {\beta}^+$ or $ {\alpha}^- \leq IOU(B^* - \delta, B^*) \leq {\beta}^-$, respectively. We accept and use it as a single input if it satisfies the IOU range for the set. We repeat the process until we have enough samples for a minibatch. Specifically, each minibatch has equal number of $B^* - \delta$ samples from $\textbf{P}^+$ and $\textbf{P}^-$.

\vspace{-0.15in}
\subsubsection{Cross-category Semi-supervised Learning}
\vspace{-0.05in}
Let $C_A$ = \{\texttt{bathtub, bed, toilet, chair, desk}\} and $C_B$ = \{\texttt{dresser, nightstand, sofa, table, bookshelf}\}.
We train with $C_{3D} = C_B$ and $C_{2D} = C_A$, i.e. we train on the 2D and 3D box labels of $C_B$ and the 2D box labels of $C_A$, to obtain the evaluation results for CS3D on $C_A$. We train with $C_{3D} = C_A$ and $C_{2D} = C_B$ to get the evaluation results for CS3D on $C_B$.

\vspace{-0.1in}
\paragraph{SUN-RGBD} 
For our 2D object detector, we train a Faster RCNN \cite{FRCNN,FRCNN_github} network on all classes $C = C_{2D} \cup C_{3D}$.
When we train the BoxPC Fit network on classes $C_{3D}$, we set the perturbation parameters to $\alpha^+ = 0.7, \beta^+ = 1.0, \alpha^- = 0.01, \beta^- = 0.25$. The loss weights for the BoxPC Fit network are set to $w_{cls} = 1, w_{reg} = 4$.
When we train the 3D object detector, we set the lower and upper bound boxes for the relaxed reprojection loss to $B^*_{lower} = B^*_{2D}, B^*_{upper} = 1.5\ B^*_{2D}$ and volume threshold for all classes to $V = 0$. The loss weights for the 3D detector are $w_{c1-reg} = 0.1, w_{c2-reg} = 0.1, w_{r-cls} = 0.1, w_{r-reg} = 2, w_{s-cls} = 0.1, w_{s-reg} = 2, w_{corner} = 0.1, w_{fit} = 0.05, w_{reproj} = 0.0005, w_{vol} = 0, w_{s-var} = 0.1$.

\vspace{-0.1in}
\paragraph{KITTI}
For our 2D object detector, we use the released detections from FPN \cite{FPN} for fair comparisons with FPN. 
We set the perturbation parameters to $\alpha^+ = 0.8, \beta^+ = 1.0, \alpha^- = 0.01, \beta^- = 0.4$
when we train the BoxPC Fit network on classes $C_{3D}$.
The loss weights of the BoxPC Fit network are set to $w_{cls} = 1, w_{reg} = 4$.
We set the lower and upper bound boxes for the relaxed reprojection loss to $B^*_{lower} = B^*_{2D}, B^*_{upper} = 1.5\ B^*_{2D}$ for \textit{Pedestrians} and \textit{Cyclists}, and $B^*_{lower} = B^*_{2D}, B^*_{upper} = 1.25\ B^*_{2D}$ for \textit{Cars}
when we train the 3D object detector on the classes $C$. We set the volume threshold for \textit{Cars} to $V^{car} = 10$ and 0 for the other classes. The loss weights for the 3D object detector are set to $w_{c1-reg} = 1, w_{c2-reg} = 1, w_{r-cls} = 1, w_{r-reg} = 20, w_{s-cls} = 1, w_{s-reg} = 20, w_{corner} = 10, w_{fit} = 1, w_{reproj} = 0.2, w_{vol} = 1, w_{s-var} = 2$.
%The experiments are performed on NVIDIA GEFORCE GTX 1080 Ti GPUs.

\vspace{-0.1in}
\subsubsection{Fully-supervised Learning}
\vspace{-0.05in}

\begin{figure*}[t]
\begin{center}
   \includegraphics[width=0.925\linewidth]{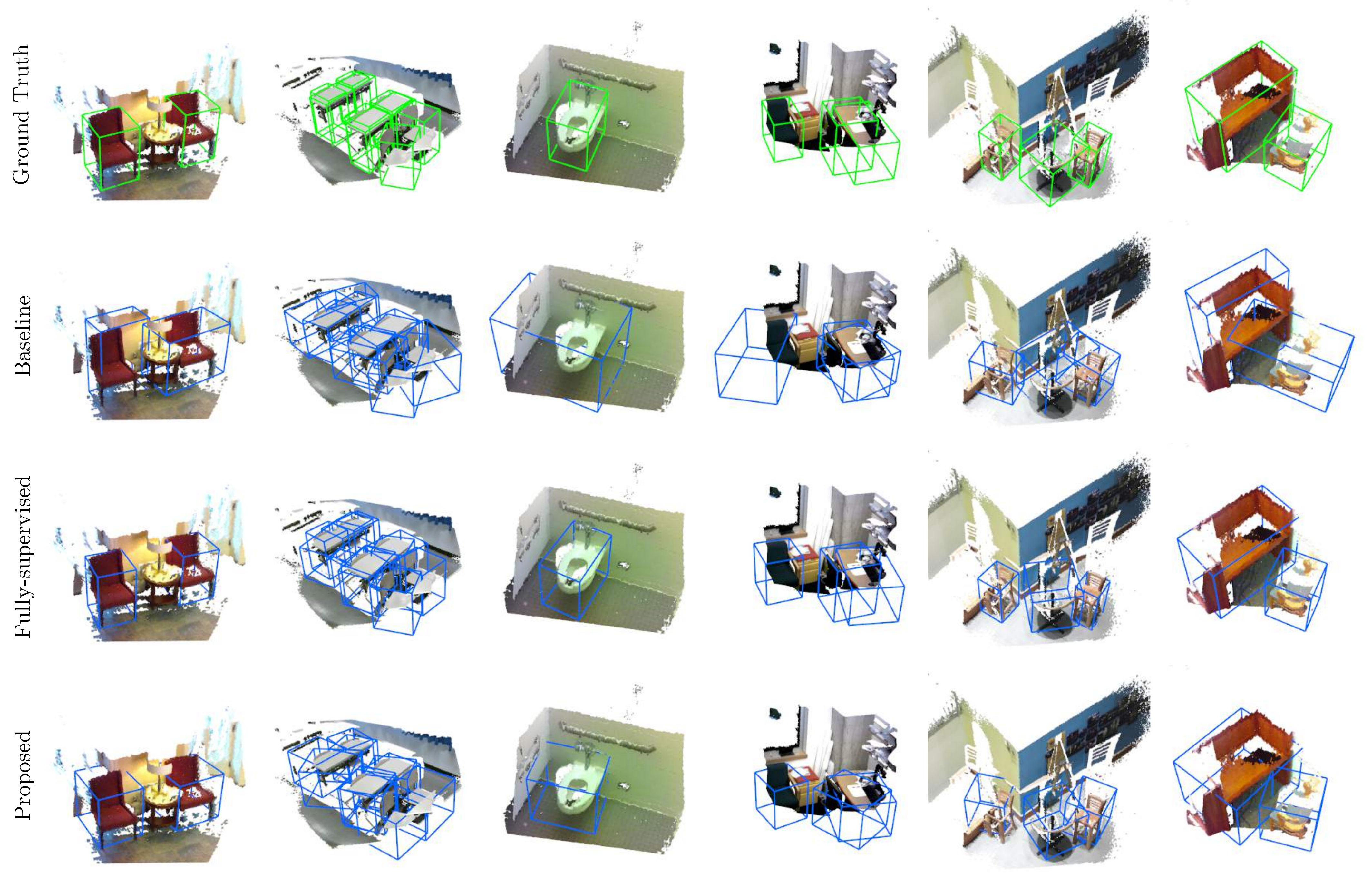}
\end{center}
\begin{center}
   \includegraphics[width=0.925\linewidth]{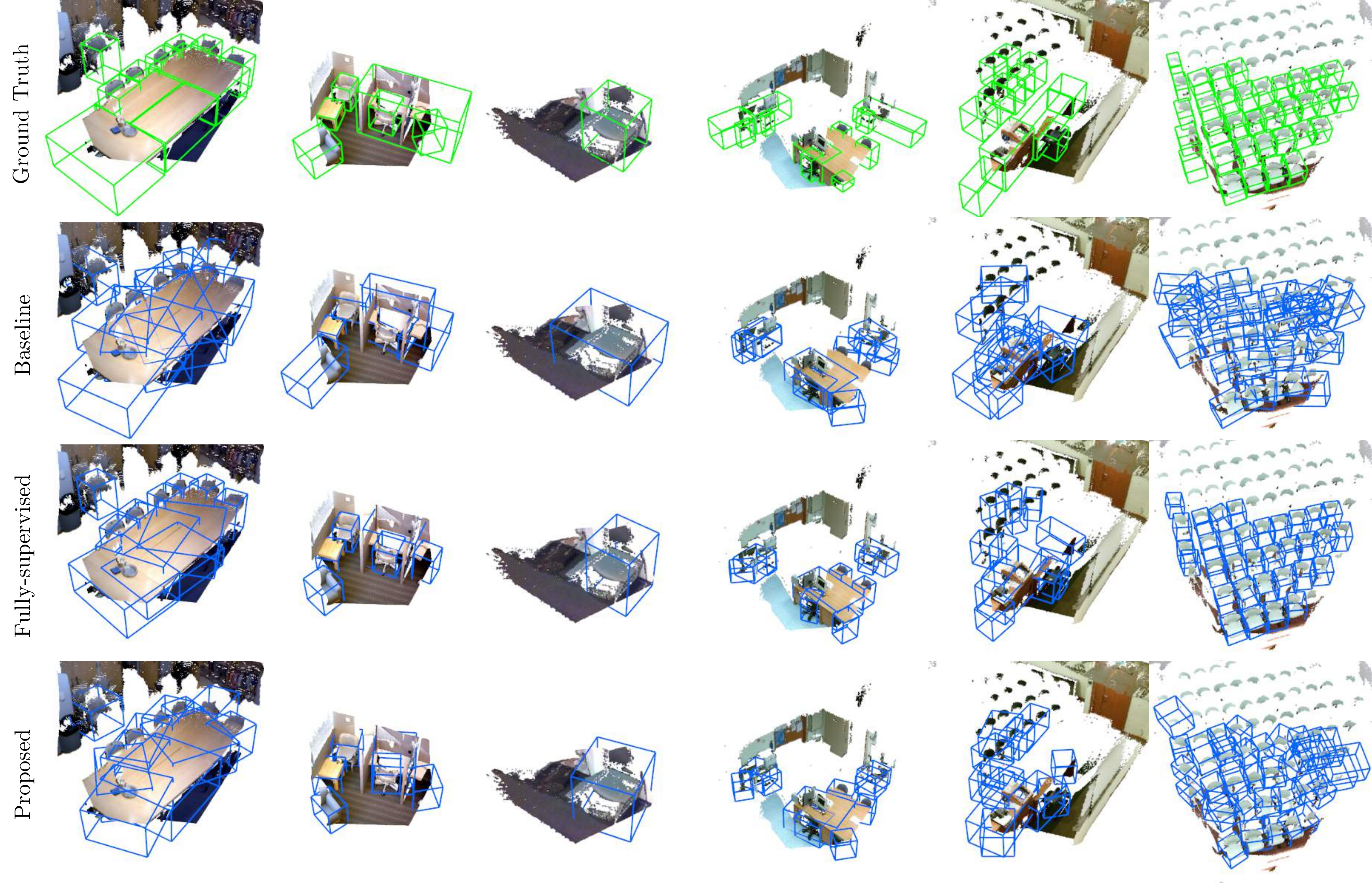}
\end{center}
   \caption{Additional qualitative comparisons between the baseline, fully-supervised and proposed models on the SUN-RGBD \textit{val} set. The baseline model's predictions tend to be large and inaccurate because it is unable to understand how to fit objects from the \textit{weak} classes. The last two examples (bottom right) are difficult scenes for the baseline and proposed models due to heavy occlusions. }
\label{fig:supp_sun_qual1}
\end{figure*}

\begin{figure*}[t]
\begin{center}
\includegraphics[width=\linewidth]{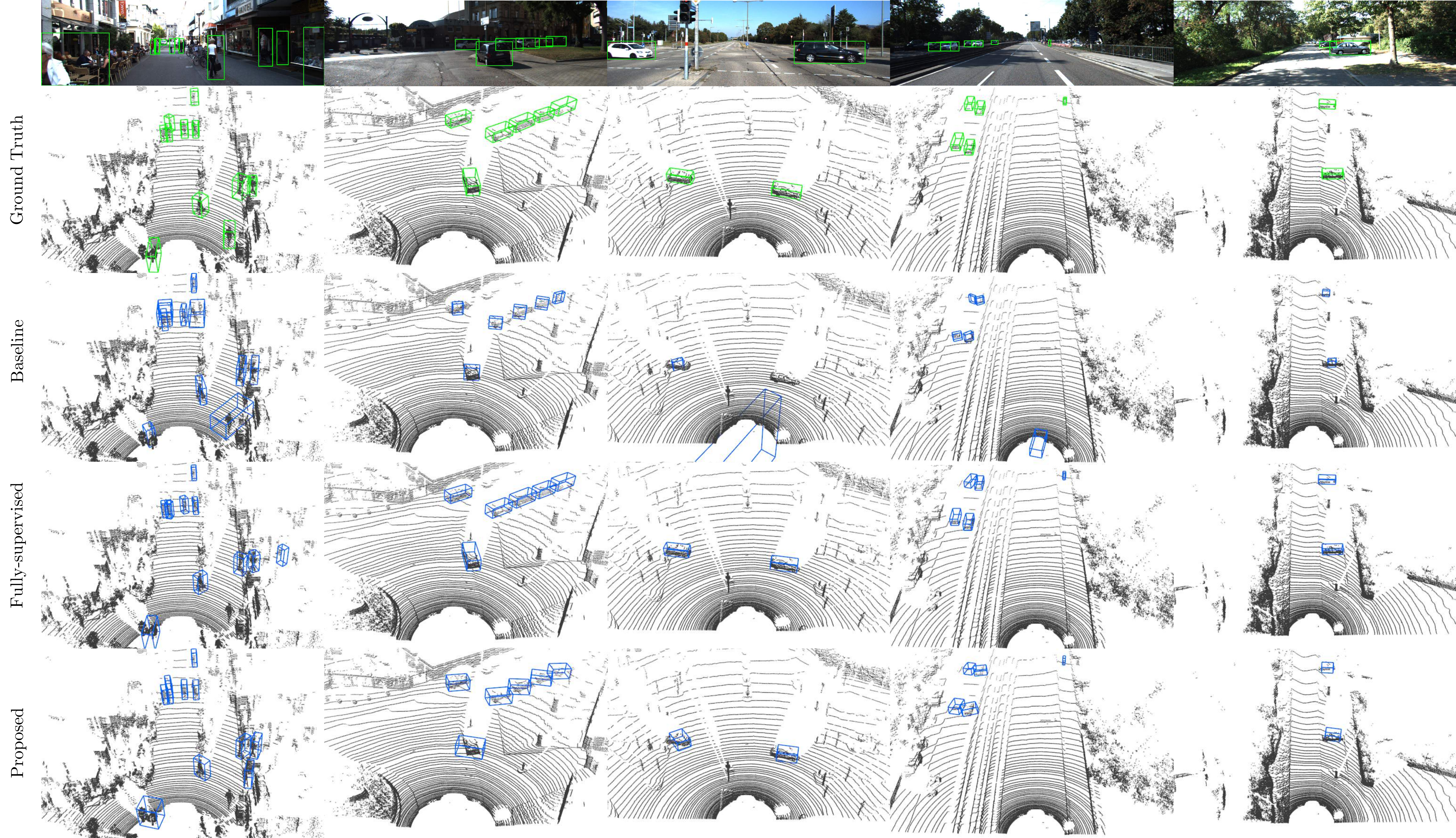}
\end{center}
\vspace{0.05in}
\begin{center}
\includegraphics[width=\linewidth]{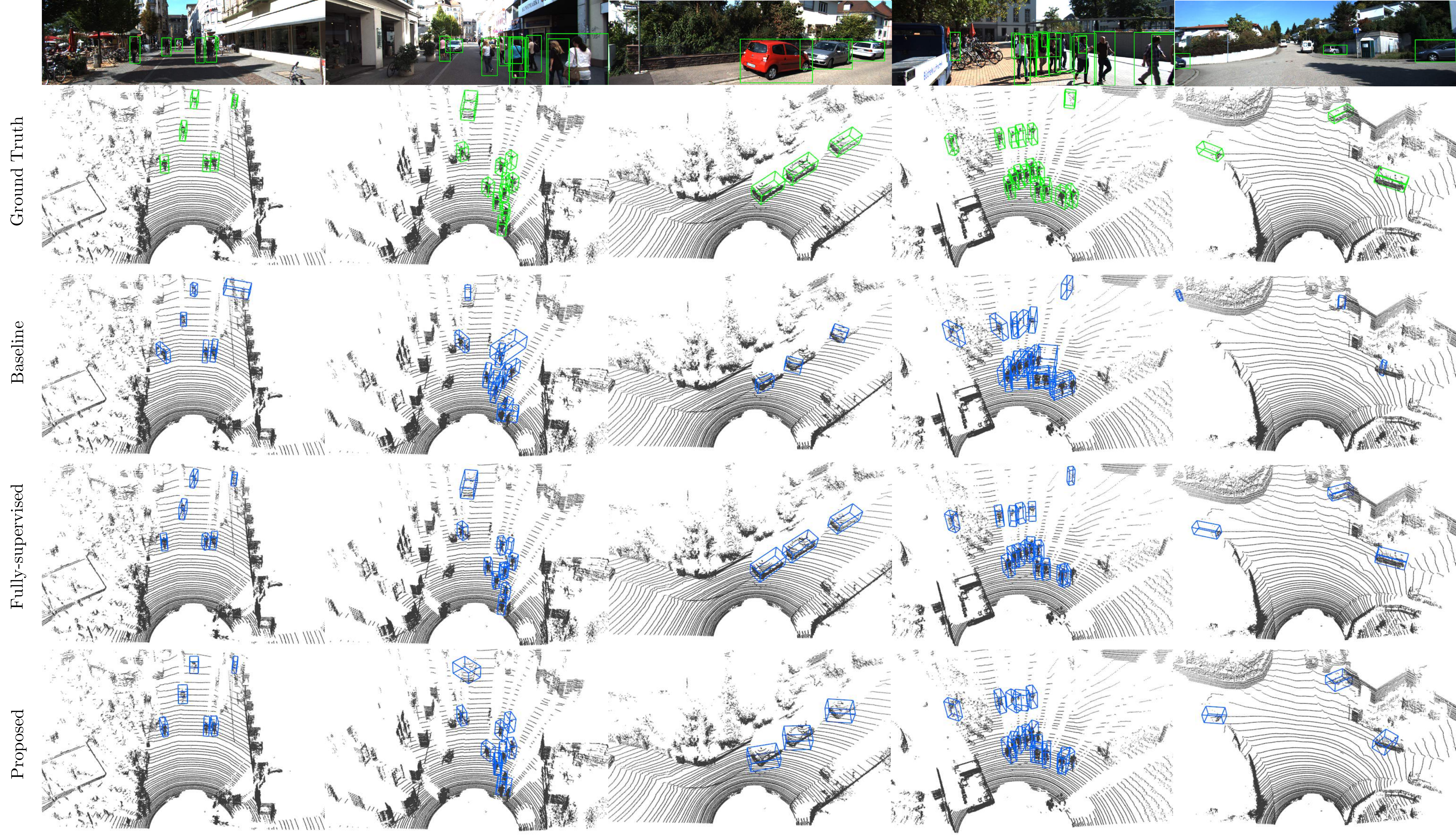}
\end{center}
   \caption{Qualitative comparisons between the baseline, fully-supervised and proposed models on the KITTI \textit{val} set. The proposed model is much closer to the fully-supervised model than the baseline model. The baseline model's predictions for \textit{Cars} tend to be inaccurate or excessively small due to the lack of a prior understanding on the scale of \textit{Cars}. }
\label{fig:supp_kitti_qual1}
\end{figure*}

\begin{figure*}[t]
\begin{center}
\includegraphics[width=\linewidth]{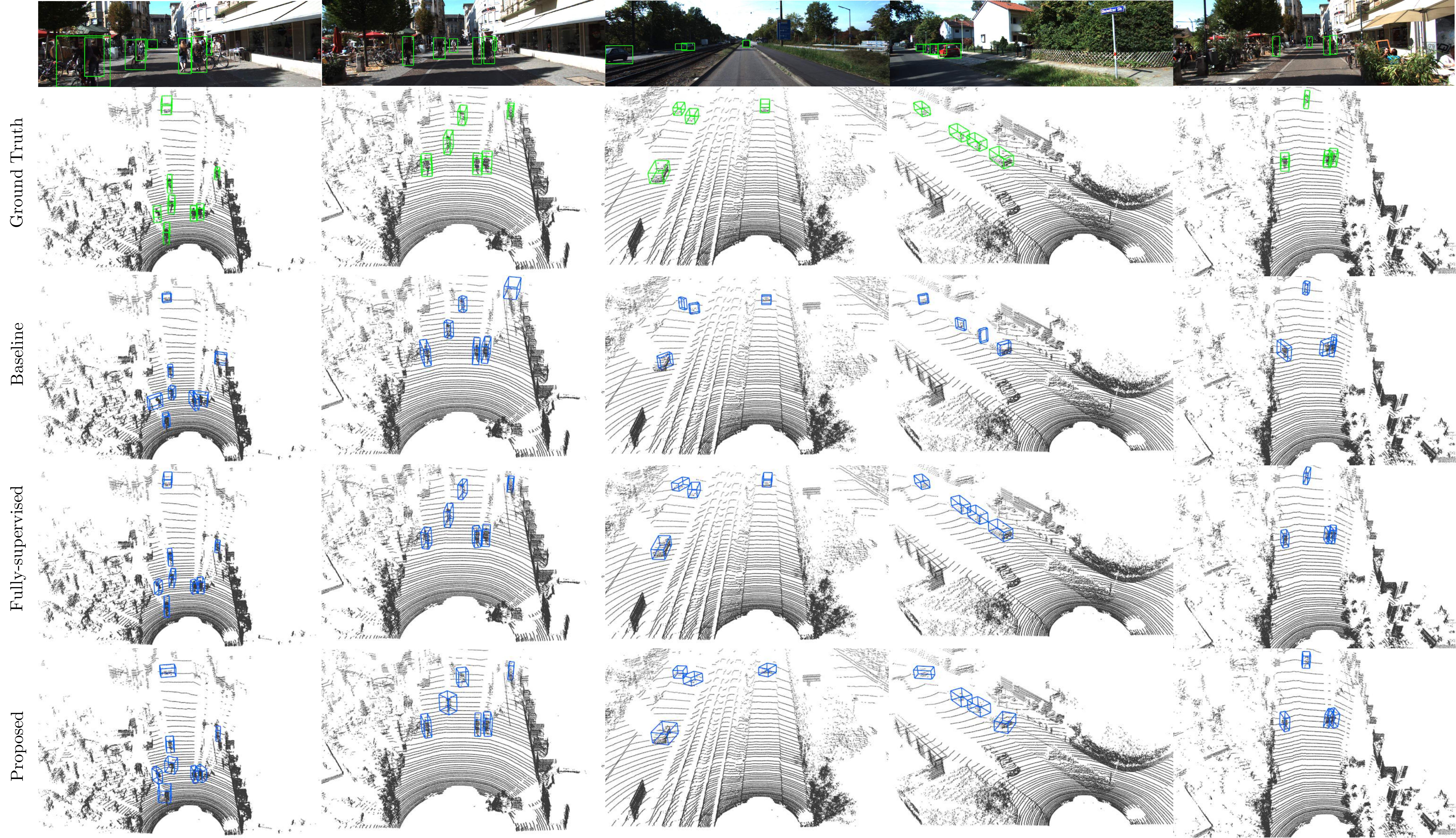}
\end{center}
\vspace{0.05in}
\begin{center}
\includegraphics[width=\linewidth]{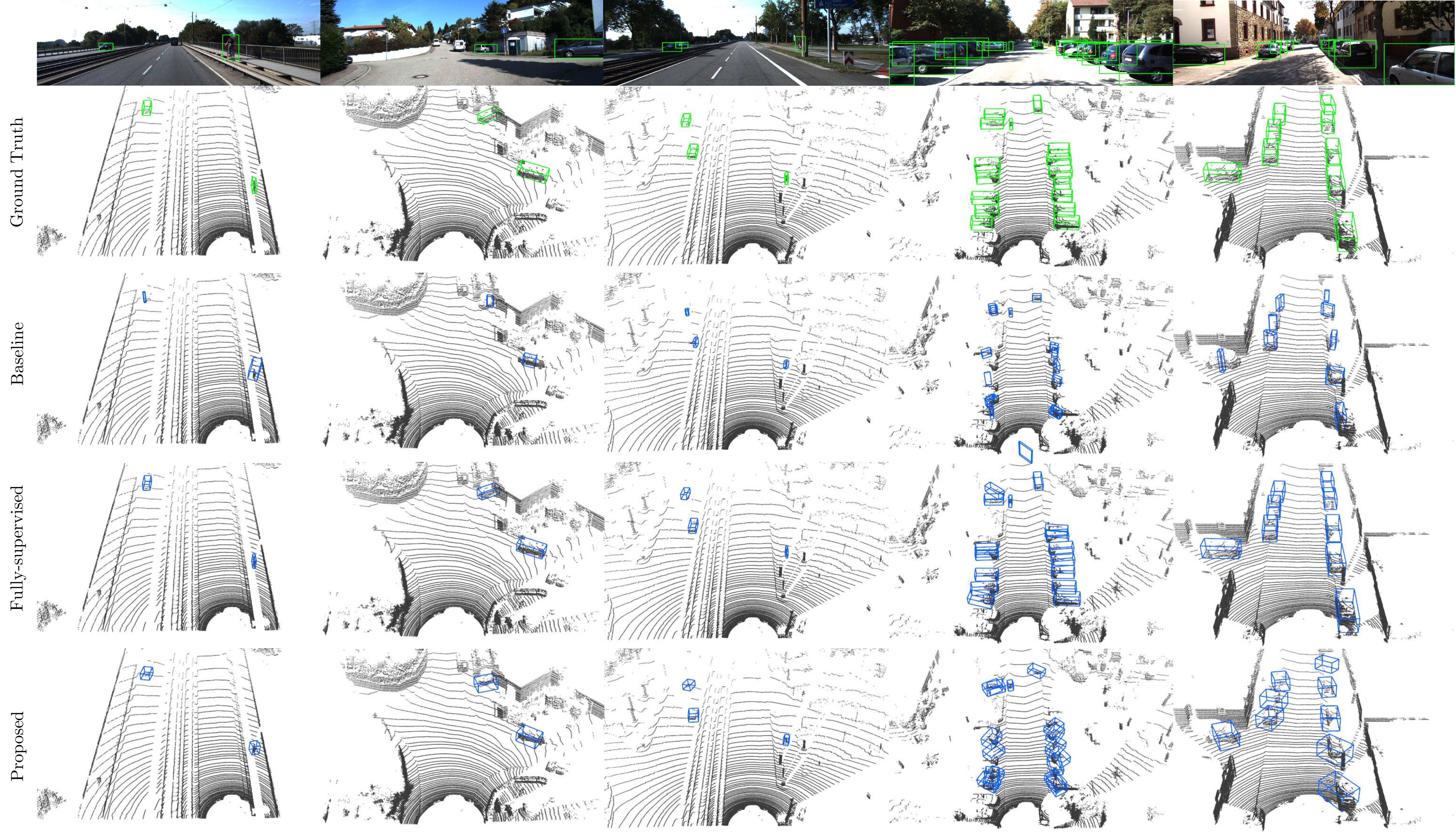}
\end{center}
   \caption{Additional qualitative comparisons between the baseline, fully-supervised and proposed models on the KITTI \textit{val} set. The last two examples (bottom right) are difficult scenes for the baseline and proposed models. This is due to heavy occlusions and poor understanding on how to fit \textit{Cars} using transferred 3D information from \textit{Pedestrians} and \textit{Cyclists}, which have very different shapes from \textit{Cars}.}
\label{fig:supp_kitti_qual2}
\end{figure*}

In the fully-supervised setting, we train and evaluate on $C$.

\vspace{-0.15in}
\paragraph{SUN-RGBD}
We use the same 2D detector as in the CS3D setting.
When we train our BoxPC Fit network, we set the perturbation parameters to $\alpha^+ = 0.6, \beta^+ = 1.0, \alpha^- = 0.05, \beta^- = 0.4$. The loss weights of the BoxPC Fit network are set to $w_{cls} = 1, w_{reg} = 4$.
For our 3D object detector, we set the parameters to $w_{c1-reg} = 0.1, w_{c2-reg} = 0.1, w_{r-cls} = 0.1, w_{r-reg} = 2, w_{s-cls} = 0.1, w_{s-reg} = 2, w_{corner} = 0.1$. Next, we use the trained BoxPC Fit network to refine the 3D box predictions of this fully-supervised model without further training.

\vspace{-0.1in}
\paragraph{KITTI}
We use the same 2D detections as the CS3D setting. 
We train one BoxPC Fit network for each of the 3 classes to allow each network to specialize on improving box predictions for a single class. For \textit{Cars}, we set $\alpha^+ = 0.7, \beta^+ = 1.0, \alpha^- = 0.6, \beta^- = 0.8$. For \textit{Pedestrians}, we set $\alpha^+ = 0.7, \beta^+ = 1.0, \alpha^- = 0.3, \beta^- = 0.7$. For \textit{Cyclists}, we set $\alpha^+ = 0.7, \beta^+ = 1.0, \alpha^- = 0.5, \beta^- = 0.7$. The loss weights of the BoxPC Fit network are set to $w_{cls} = 0, w_{reg} = 4$. Next, we use the trained BoxPC Fit networks to refine the 3D box predictions of the 3D object detector model released by \cite{FPN} without further training.

\subsection{Additional Results for SUN-RGBD} \label{supp_sun}
In Fig.~\ref{fig:supp_sun_pr_curves}, we plot the precision-recall curves for different methods to study the importance of each proposed component. 
All the methods are cross-category semi-supervised except for ``FPN (Fully-supervised)'', which is the fully-supervised FPN \cite{FPN} that gives an upper-bound performance for the methods. We observe that ``Ours + BoxPC + R + P'' (in yellow) has higher precision at every recall than the baseline ``FPN w/o OneHot'' (in blue) for almost all classes.

We also provide additional qualitative results on the SUN-RGBD dataset in Fig.~\ref{fig:supp_sun_qual1}. The predictions made by the baseline model tend to be large and inaccurate due to the lack of strong labels in the \textit{weak} classes. In the third example of Fig.~\ref{fig:supp_sun_qual1}, we see that the predictions by the baseline model can also be highly unnatural as it cuts through the wall. On the contrary, we observe that the proposed model's predictions tend to be more reasonable and closer to the fully-supervised model despite not having strong labels for the \textit{weak} classes. In the last two examples of Fig.~\ref{fig:supp_sun_qual1}, the heavy occlusions in the scene makes it difficult for the baseline and proposed models to make good predictions.

\subsection{Qualitative Results for KITTI} \label{supp_kitti}
In Fig.~\ref{fig:supp_kitti_qual1} and Fig.~\ref{fig:supp_kitti_qual2}, we provide qualitative comparisons between the baseline, fully-supervised and proposed models for the KITTI dataset. In both Fig.~\ref{fig:supp_kitti_qual1} and Fig.~\ref{fig:supp_kitti_qual2}, we again observe that the proposed model is closer to the fully-supervised model than the baseline model. Notably, in the first example (top left) and ninth example (bottom right) of Fig.~\ref{fig:supp_kitti_qual1}, we observe that the proposed model is able to make significantly good predictions on \textit{Pedestrians} despite the crowded scene. 

For almost all of the \textit{Car} instances, the baseline model makes excessively small predictions because it was trained on \textit{Pedestrian} and \textit{Cyclist} classes which are much smaller. Our proposed model is able to make better predictions on \textit{Cars} but the orientation of the 3D box predictions can be improved further.

In the last two examples of Fig.~\ref{fig:supp_kitti_qual2}, we show some of the difficult scenes for our baseline and proposed models. This is because there are heavy occlusions and poor understanding of \textit{Cars} since the models were trained on \textit{Pedestrians} and \textit{Cyclists}, which have very different shapes and sizes from \textit{Cars}.

\end{document}